%% file: main.tex
\documentclass{ieeeconf}
\usepackage[utf8]{inputenc}
\usepackage{graphicx}
\usepackage{subcaption}
\usepackage{hyperref}
\usepackage[capitalise]{cleveref}
\usepackage{caption}
\newcommand{\argmax}{\arg\!\max} 
\usepackage[top=54pt, left=54pt, right=54pt, bottom=54pt]{geometry}
\title{Adaptive Sampling using POMDPs with Domain-Specific Considerations} %

\author{Gautam Salhotra\textsuperscript{\textdagger}, Christopher E. Denniston\textsuperscript{\textdagger}, David A. Caron, Gaurav S. Sukhatme\textsuperscript{\textdaggerdbl}%
\thanks{\textsuperscript{\textdagger} Equal contribution.{\tt\small \{salhotra, cdennist\}@usc.edu}}\thanks{All authors are with the University of Southern California}\thanks{\textsuperscript{\textdaggerdbl} G.S. Sukhatme holds concurrent appointments as a Professor at USC and as an Amazon Scholar. This paper describes work performed at USC and is not associated with Amazon.}
\thanks{This work was supported in part by the Southern California Coastal Water Research Project Authority under prime funding from the California State Water Resources Control Board on agreement number 19-003-150 and in part by USDA/NIFA award 2017-67007-26154.}}%

\usepackage{subfiles} %

\begin{document}

\maketitle

\begin{abstract}
\subfile{sections/0_abstract}
\end{abstract}

\section{Introduction}
\subfile{sections/1_intro}

\section{Background}
\subfile{sections/3_background}

\section{Formulation and Approach}
\subfile{sections/4_formulation}

\section{Experiments}
\subfile{sections/5_experiments}

\section{Conclusion}

\subfile{sections/6_conclusion}

\bibliographystyle{unsrt}
\bibliography{references}

\clearpage
\section{Appendix}
\subfile{sections/8_appendix}
\subsection{Future Work}
\subfile{sections/7_future_work}

\end{document}

%% file: sections/0_abstract.tex
We investigate improving Monte Carlo Tree Search based solvers for Partially Observable Markov Decision Processes (POMDPs), when applied to adaptive sampling problems.
We propose improvements in rollout allocation, the action exploration algorithm, and plan commitment. The first allocates a different number of rollouts depending on how many actions the agent has taken in an episode. We find that rollouts are more valuable after some initial information is gained about the environment. Thus, a linear increase in the number of rollouts, i.e. allocating a fixed number at each step, is not appropriate for adaptive sampling tasks. The second alters which actions the agent chooses to explore when building the planning tree. We find that by using knowledge of the number of rollouts allocated, the agent can more effectively choose actions to explore. The third improvement is in determining how many actions the agent should take from one plan. Typically, an agent will plan to take the first action from the planning tree and then call the planner again from the new state. Using statistical techniques, we show that it is possible to greatly reduce the number of rollouts by increasing the number of actions taken from a single planning tree without affecting the agent's final reward. Finally, we demonstrate experimentally, on simulated and real aquatic data from an underwater robot, that these improvements can be combined, leading to better adaptive sampling. The code for this work is available at \url{https://github.com/uscresl/AdaptiveSamplingPOMCP}.

%% file: sections/1_intro.tex
Adaptive sampling is the process of intelligently sampling the environment by an agent, such as an underwater or aerial robot. The robot does this by creating an internal model and selecting sampling positions that improve the model~\cite{hwang_auv_2019}.
Adaptive sampling is often preferred to full workspace coverage plans when either a. the robot cannot cover the entire workspace due to a constrained time or energy budget, or b. an approximate model of the workspace is acceptable. 
Adaptive sampling can also make use of domain-specific information when it is available.  
For example, researchers in the area of algal bloom monitoring in aquatic ecosystems find areas of high chlorophyll concentration more valuable to study. Such use-cases naturally lend themselves to the integration of Bayesian optimization in adaptive sampling~\cite{das_towards_2010}.

Solving adaptive sampling problems exactly is known to be NP-hard~\cite{krause_submodularity_2011}. 
However, these problems can be solved using exact solvers~\cite{binney_branch_2012}, sampling-based planners \cite{Hollinger2014} or Monte Carlo tree search (MCTS)-based solvers which sample random trajectories from the final reward distribution~\cite{Marchanta,best_dec-mcts_2019}.
Here we focus on using MCTS-based solvers to effectively sample complex environments.
These iterative solvers generally use rollouts to sample a reward value for a given state by following trajectories from that state. 
The process of sampling from a final reward distribution using a random trajectory from a state at the leaf of the planning tree is called a rollout. 
Rollouts are used in local planners, such as POMCP~\cite{silver_monte-carlo_2010}, to sample discounted rewards over trajectories, from an unknown reward distribution.

\begin{figure}[t]
    \centering
    \subcaptionbox{\label{fig:drone_baseline}}{\includegraphics[height=3cm]{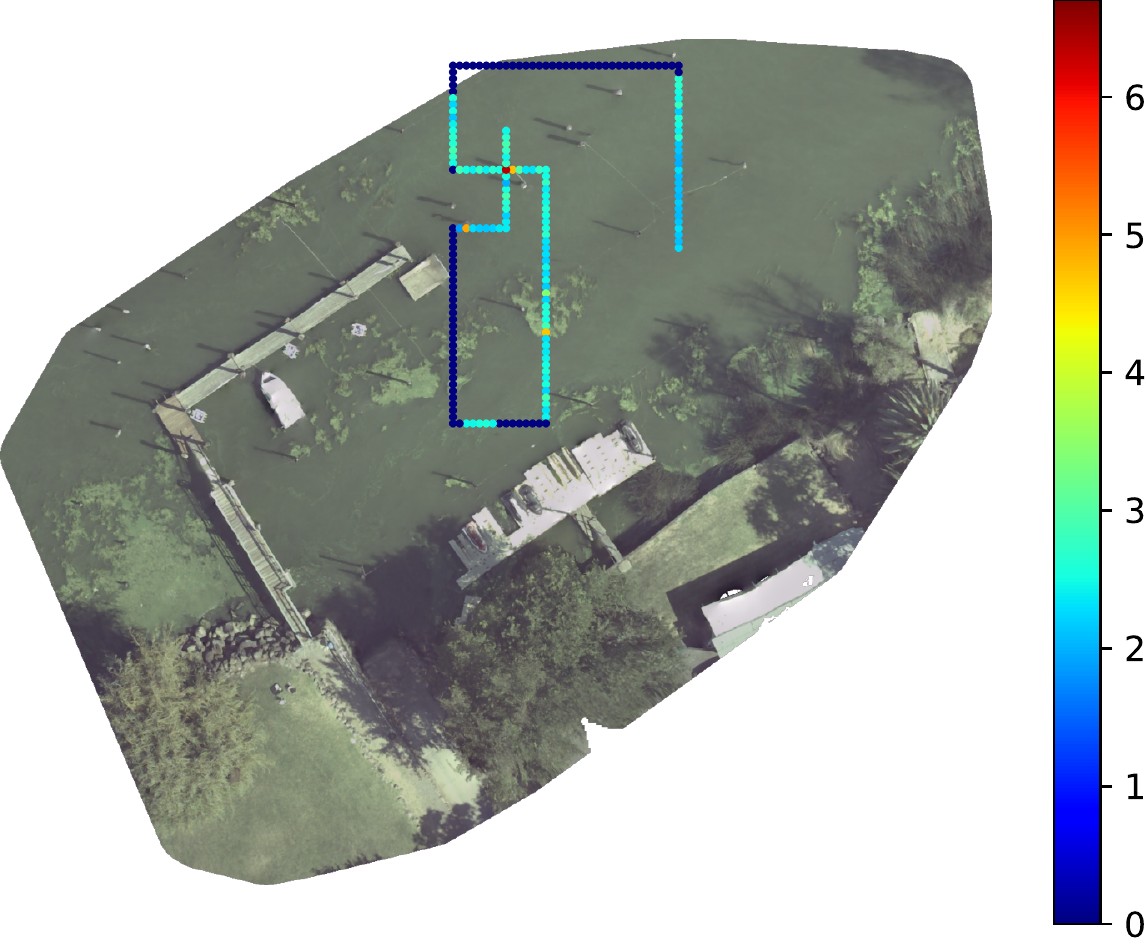}}

    \subcaptionbox{\label{fig:block_diagram}}{\includegraphics[width=.45\textwidth]{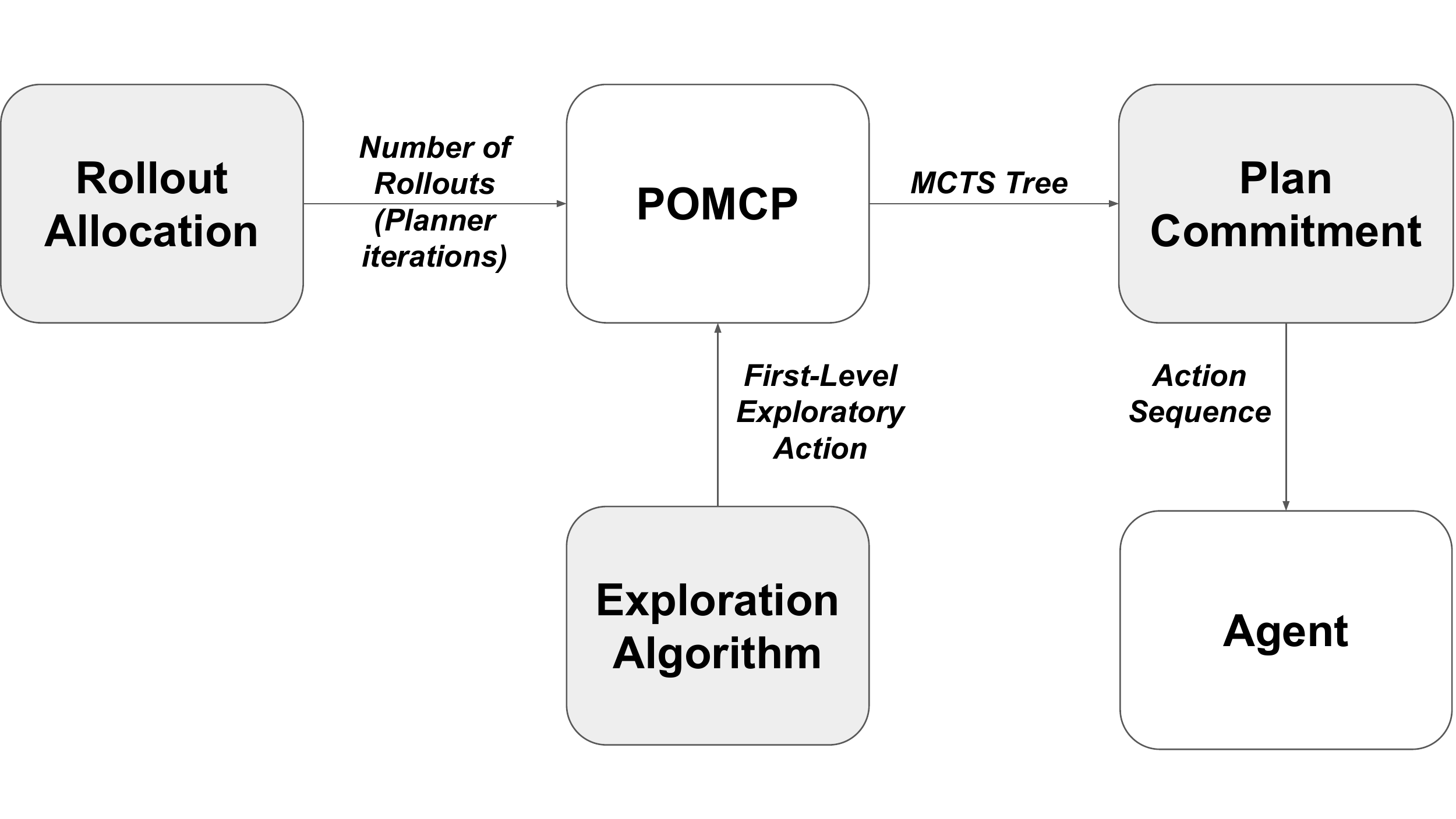}}

    \caption{Fig.~\subref{fig:drone_baseline} shows an agent's trajectories on a hyperspectral orthomosaic collected in Clearlake, California. Blue is low value, red is high value. The baseline trajectory overlaps with itself (wasted time), goes outside the bounds of the orthomosaic (no reward outside workspace), and mostly samples near the starting position. The trajectory from our proposed method avoids such behavior and samples regions further away. Fig.~\subref{fig:block_diagram} shows POMDP planning with the POMCP planner. Portions in grey are areas we study and improve in this work.}
\vspace{-0.25in}
\end{figure}

In adaptive sampling, this reward distribution is defined by some objective function over samples. %
Typically, rollouts are used to build an estimate of the mean reward for an action.
By performing more rollouts, the planner improves its estimate of the expected reward for a particular sequence of actions.
Often, planning for adaptive sampling is done online. A fixed number of environment steps (typically one) are enacted after planning for a fixed number of iterations.
This process is repeated until the finite budget (e.g., path length or energy~\cite{Hollinger2014}) is exhausted. 
Here, we verify that this process of committing to a fixed number of steps and rollouts at each invocation of the planner can be modified to reduce the total number of rollouts needed over the entire episode.
We find that, in information gathering problems, there is a period when the information gathered is sufficient to predict the field accurately enough to make more useful plans. 
The intuition behind our result is that this period should be allocated more rollouts than the period when less information is known, or when gathering more samples does not result in as  much reward.
Additionally, more environment steps can be enacted from a single POMCP planning tree because the reward for future actions can be accurately predicted.

We cast the adaptive sampling problem as a Partially Observable Markov Decision Process (POMDP) which is solved using a local solver that updates the expected rewards for action sequences  by sampling some reward distribution in an iterative fashion using rollouts.
Specifically, we investigate minimizing the number of rollouts performed to achieve comparable accumulated reward by altering three parameters: the number of rollouts to perform, the choice of which actions to take in the planning tree during a planning iteration, and how many steps of the planned trajectory to follow.
\cref{fig:drone_baseline} shows a sample trajectory of a drone for a lake dataset performed by our method and the baseline POMCP solver in this work.

%% file: sections/3_background.tex
\textbf{Gaussian Processes} are widely used modeling tools for adaptive sampling because of their non-parametric and continuous representation of the sensed quantity with uncertainty quantification~\cite{Hollinger2014,kemna_pilot_2018,Marchanta}.
Gaussian processes approximate an unknown function from its known outputs by computing the similarity between points from a kernel function~\cite{Rasmussen2006}.
Gaussian Processes are specifically useful for modeling the belief distribution of the underlying function from observations in the POMDP formulation of Bayesian optimization~\cite{borenstein_bayesian_2014}.

\textbf{Online Adaptive Sampling} consists of constructing an optimal path by alternating between planning and action. 
A plan is developed which attempts to maximize some objective function $f$ by taking the actions describe by partial trajectory $p$.
The partial trajectory is executed and samples are added to the model of the environment. 
These partial trajectories are concatenated to form full trajectory $P$.
The plan and act iterations are iteratively interleaved until the cost $c(P)$ exceeds some budget $B$. 
Formally this is described by \cref{eq:problem_statement}.
\begin{equation}\label{eq:problem_statement}
    P^* = \argmax_{P \in \Phi} f(P) | c(P) \leq B
\end{equation}
where $\Phi$ is the space of full trajectories, and $P^*$ is the optimal trajectory~\cite{Hollinger2014}.

Typically, $f$ is an objective function describing the quality of the model of the environment sampled by $P$ is. In this work, the objective function is $\mu_x + c \sigma_x$, where $\mu_x$ is the Gaussian process estimate of the underlying function value at $x$ and $\sigma^2_x$ is the variance of the Gaussian process estimate at $x$. 
This objective function is commonly used in Bayesian adaptive sampling with a $c$ parameter to trade off between exploration and exploitation.
The exploration term of the objective function, $\sigma(x)$, exhibits submodularity characteristics~\cite{guestrin_near-optimal_2005}.
Formally a function, $F$, is submodular if $\forall A \subset B \subset V$ and $\forall s \in V \setminus B$, we have that $F(\{s\} \cup A) - F(A) \geq F(\{s\} \cup B) - F(B)$~\cite{krause_submodularity_2011}. 
This naturally describes diminishing returns exhibited in many adaptive sampling problems where taking a sample provides more information if you have taken fewer samples before this.

\textbf{Partially Observable Markov Decision Processes (POMDPs)} are a framework for solving estimation problems when observations do not fully describe the state. 
It has been shown that formulating observations as samples from the underlying world state and representing the robot's model of the underlying function as a belief state can be formulated as a Bayesian Search Game~\cite{borenstein_bayesian_2014}, a framework for solving Bayesian optimization problems.
In this game, an agent has to select points in the domain $X$ that maximize the value of an unknown $g(x)$. 
Samples from $g(x)$ constitute observations which are partially observable components of the overall state $g$.
If this state is augmented with the state of the robot, $x$, and constrained to locally feasible robot actions, this formulation can easily be extended to adaptive sampling~\cite{Marchanta}.
To represent the underlying belief $b$ at each state, a Gaussian process may be used.
We formulate the adaptive sampling problem as a POMDP as shown in \cref{tab:BayesianSearchAndPOMDPs}.

\begin{table}[]
\begin{tabular}{c|c}
\textbf{POMDP} & \textbf{Adaptive Sampling}        \\ \hline
States         & Robot position $x_t$, Underlying unknown function $g$ \\ \hline
Actions        & Neighboring search point $x_{t+1}$                    \\ \hline
Observations   & Robot position $x_t$, Sampled value $y_t = g(x_t)$    \\ \hline
Belief         & Gaussian Process $GP(y_{1:t}| x_{1:t})$                                \\ \hline
Rewards        & $f(x_t) = \mu_{x_t} + c * \sigma_{x_t}$                       
\end{tabular}
\caption{Adaptive Sampling as a POMDP \cite{borenstein_bayesian_2014}}
\label{tab:BayesianSearchAndPOMDPs}
\vspace{-0.25in}
\end{table}

\textbf{Partially Observable Monte Carlo Planning (POMCP)}: 
POMDPs have been used for adaptive sampling and informative path planners in many situations~\cite{lermusiaux_science_2016,kim2021plgrim,Marchanta,hai-feng_underwater_2019}.
Many of these use a traditional dynamic programming solution to solve the underlying POMDP.
This is infeasible for large state spaces or complex belief representations that are typically present in many common adaptive sampling problems.
Recently, attention has focused to solvers which are locally accurate using probabilistic rollout updates~\cite{Marchanta}.
A state of the art algorithm for solving large POMDPs online is the Partially Observable Monte-Carlo Planning solver (POMCP)~\cite{silver_monte-carlo_2010}.
POMCP uses a Monte Carlo tree search (MCTS) which propagates reward information from simulated rollouts in the environment. 
At every iteration in the planner, the agent performs a search through the generated tree $\mathcal{T}$ to choose actions, using Upper-Confidence Tree (UCT)~\cite{UCT_kocsis2006bandit} exploration for partially observable environments, until it reaches a leaf node $\mathcal{L}$ of $\mathcal{T}$.
From node $\mathcal{L}$, the planner performs a rollout using a pre-defined policy (usually a random policy) until the agent reaches the planning horizon.
The reward the agent collects while simulating this trajectory is then backpropagated up through the visited states. 
Finally, the first node the agent visited from the rollout is added as a child of $\mathcal{L}$, and the tree is expanded. 
Once the specified number of iterations (rollouts) are completed, the tree is considered adequate for planning. 
The action from the root node that gives the highest expected reward is then chosen, and the agent executes the action in the environment.
At each observation node of $\mathcal{T}$, the observation is estimated with $\mu_x$, where $x$ is the agent's position at that node.
To update the belief $b$, the state-observation pair is integrated into the Gaussian Process. %

\textbf{Multi-Armed Bandits (MAB)} are a family of problems in which $K$ actions are available and each action has an unknown associated reward distribution, $D_k$.
At each time step, the agent chooses an action $k \in K$ and receives a reward drawn from $D_k$.
The goal of the agent is to maximize the cumulative reward over time or minimize risk, which is defined as the difference between the agent's cumulative reward and some optimal policy. 
There is a natural exploration-exploitation trade-off in the MAB problem because at each step the agent receives more information about the distribution of one of the actions by sampling the reward~\cite{bouneffouf_survey_2020}.
This framework provides the mechanism for action selection in a variety of rollout-based algorithms, including POMCP~\cite{silver_monte-carlo_2010}, and is used when each rollout can be viewed as a draw from the reward distribution conditioned on the currently selected actions.

In contrast to optimal action selection algorithms, there is a family of algorithms which seek to identify the best action in terms of mean reward.
These algorithms work in two settings: fixed-budget and fixed-confidence. 
In the fixed-budget setting, the agent finds the best action from a fixed number of samples. In the fixed-confidence setting, the agent finds the best arm having $P[risk > \epsilon] < \delta$, in the fewest number of samples~\cite{gabillon_best_2012}.

%% file: sections/4_formulation.tex
Most online informative planning pathfinders use the same planning duration at all points in the planning sequence. We propose to modify the POMCP~\cite{silver_monte-carlo_2010} planner, an online, rollout-based POMDP-planner, given the knowledge about the underlying problem.
Our method selects how many rollout iterations to use at each environment interaction and which actions should be tried in the planner.
Using this tree, it also adaptively selects how much of the tree can be incorporated into the executed plan, based on the rewards received during these rollouts.
We show that we can produce similar models of the environment in fewer overall iterations of the (expensive) rollout sequence.
An overview of how our improvements fit into the planning and action pipeline with POMCP is in~\cref{fig:block_diagram}.

\vspace{-0.1in}
\subsection{Rollout Allocation}
\label{rolloutAllocation}
The first improvement we propose is to alter how the overall rollout budget is handled.
Typically, the total rollout budget is divided evenly and a fixed number of rollouts are allocated each time the planner is called to compute a new partial trajectory.
This results in a linear increase in the number of rollouts used as the number of environment steps increases.
We propose that this rollout allocation method should take advantage of three key ideas: cold-starting, submodularity, and starvation. 
The idea of cold-starting is well studied in adaptive sampling~\cite{kemna_pilot_2018} and captures the notion  that the planner can not make useful decisions with little information to plan on. 
Planning with little information is futile since far away points will generally return to the global mean with high variance.
Typically, this is handled by having the robot perform a pre-determined pilot survey to gather initial information about the workspace~\cite{kemna_pilot_2018}.
This strategy wastes sampling time if too many pre-programmed samples are taken. 
The lack of information manifests itself in another problem when planning for adaptive sampling: submodularity of the objective function which necessitates effective sampling early on in the episode, since early samples provide the largest benefit.
Additionally, because this information is used in planning later on, the importance of good early information gathering is compounded.
There is, of course, a trade-off to allocating rollouts early on: plans which allocate too many rollouts early on could suffer from the problem of starvation of rollouts at later stages.
This can cause the planner to make poor decisions when there is rich information to plan on and the areas of interest are well-known and understood.
This rollout allocation trade-off in planning is an instance of the exploration-exploitation trade-off which is common in information gathering tasks.

\subsection{Exploration Algorithm}
\label{armSelection}
MCTS-based planners, such as POMCP, treat each action at each layer in the tree as an MAB problem.
In the usual MCTS algorithm, the objective is to optimally explore possible trajectories by choosing an action at each layer according to some optimal action selection criterion~\cite{silver_monte-carlo_2010}.
We propose using optimal arm identification algorithms instead of optimal exploration algorithms because the final goal of the POMCP is to choose and execute the action with the highest mean reward, not to maximize the cumulative reward during searching. 

Shifting from the cumulative-reward setting to the fixed-budget setting allows the exploration algorithm to decide the exploration-exploitation trade-off based on the number of allocated rollouts at each planning step.
When many rollouts are allowed, the algorithm can be more explorative and consider each action for longer, while with fewer rollouts the algorithm typically becomes more exploitative.
A fixed-budget action selection algorithm can only be used for the first action in the tree as these actions are the only ones for which the total number of combined rollouts is fixed. 

In this work, we investigate three exploration algorithms.
The first, Upper Confidence Tree (UCT), is an optimal exploration algorithm and is the default choice for most MCTS-based solvers, including POMCP~\cite{silver_monte-carlo_2010}. 
UCT provides a trade-off between exploration and exploitation by adding an exploration bonus to each action using the number of  times the parent and child have been explored.
UCT does not take into account a budget, but instead tries to maximize the sum of the reward samples during planning. 
Because of this, UCT may tend to be highly explorative~\cite{audibert_best_2010}.

The remaining two algorithms are fixed-budget algorithms that explicitly incorporate the amount of rollouts allotted and attempt to maximize the probability that the chosen action is, in fact, the best action.
The first algorithm, UGapEb, uses an upper bound on the simple regret of the actions to choose the action which is most likely to switch to become the best one to exploit~\cite{gabillon_best_2012}. 
This algorithm incorporates the total budget into the confidence interval to make the most efficient use of the provided budget.
It contains a (difficult to estimate) parameter, $H_\epsilon$, which requires the gap between actions to be known ahead of time. 

Both UCT and UGapEb additionally depend on a (difficult to estimate) parameter, $b$, which is multiplied by the bound to create the exploration-exploitation trade-off.
In this work, we use the difference between the highest and lowest discounted rewards ever seen, but this may not be the optimal bound and requires the agent to know these values before exploring the workspace.

The final exploration algorithm is Successive Rejects. 
This algorithm is unique in that it does not attempt to use a confidence bound like UCT and UGapEb. Instead, it chooses each action a number of times in successive rounds and eliminates an action in each round until a single action is found~\cite{audibert_best_2010}. 
This algorithm is preferable in some situations because it is parameter-free, while UCT and UGapEb have hard to tune parameters. However it may waste some rollouts when an action is obviously inferior.

\subsection{Plan Commitment}
\label{planCommitment}
Each call to the POMCP planner produces a planning tree of estimated rewards for action sequences.
Typically, the agent executes the first action with the highest expected reward and replans with the part of the tree it went down.
Since an adaptive sampling agent is approximating the underlying function using an initial belief, the generative model is different after each call to POMCP planner.
This is because every time the agent takes an action in the environment and receives new samples, the initial belief for the underlying POMDP changes. 
Hence, the tree must be discarded and completely replanned after incorporating the observation from the environment into the initial belief state.
We propose to take more than one action from the tree when there is certainty at lower levels about the optimality of the action.
In order for the agent's performance to be unaffected by taking further actions, there are two considerations it must have.
The first consideration is of the quality of the estimate of the reward from an action.
If rollouts are spread evenly, the number of rollouts performed for lower actions (deeper in the tree) will be exponentially less than higher actions (closer to the root), causing the estimate of their reward to be poor.
The second consideration is the quality of the estimate of the observation at locations further away.
The initial belief greatly affects the quality of observation estimates.  Trajectories further away from what the agent has seen will generally have worse estimates for what their observations will likely be.

With these two considerations the agent may be able to extract more than one plan step from the tree without significant deterioration in the accumulated reward.

The simplest method is to take a fixed number of actions from the MCTS and execute them.
This does not take into account any considerations of the quality of estimates of the reward for actions. 
With this method,  the agent may have an inadequate understanding of the underlying world state at some point, but still take actions from the tree.

If the agent accounts for the statistics of the samples for the reward for actions, more complex methods can be used.
We use a method based on UGapEc, a fixed-confidence MAB scheme~\cite{gabillon_best_2012}.
UGapEc determines if there is an action that has a mean reward higher than all other actions with probability $1-\delta$. 
This  algorithm  can  be  used  to  verify  if  a  fixed-confidence  threshold  is  met  by  checking  if  the  algorithm would not choose to explore further.

Another method which uses a statistical test similar to UGapEc is a two tailed Welch's t-test~\cite{welch_generalisation_1947}.
This test assumes the distributions the samples are from are Gaussian but does not assume the standard deviations are known or equal. 
Since the error in the estimate of standard deviation is quadratic in the number of samples, our estimate of the standard deviation deteriorates much faster than our estimate of the mean.
Because of this, a more complex test must be used than a simple Gaussian confidence interval test since it may underestimate the sample standard deviation~\cite{gurland_simple_1971}.
The unequal variances two tailed t-test tests the null hypothesis that the reward distributions for two actions have identical expected values.
This method returns an estimate of the probability (p-value) that the means are equal. 
A threshold is set and if the p-value of the null hypothesis is below the threshold, the action is considered safe to take. 
This method is statistically robust. It causes the action to not be chosen in two cases.
The first case is that there are not enough samples of each action and the second is that the means are too close to distinguish with the number of samples gathered.
This method uses a Student's t-distribution~\cite{gosset1908_students_tdist} and calculates the t statistic and the v value with \cref{welchvalue} which can be used to compute the p-value with a Student's t-distribution.

\vspace{-0.2in}
\begin{equation}\label{welchvalue}
    t = \frac{\bar{\mu}_1 - \bar{\mu}_2}{\sqrt{\frac{\bar{\sigma}^2_1}{N_1} + \frac{\bar{\sigma}^2_2}{N_2}}}~~~~~v \approx \frac{(\frac{\bar{\sigma}^2_1}{N_1} + \frac{\bar{\sigma}^2_2}{N_2})^2}{\frac{\bar{\sigma}^4_1}{N^2_1(N_1-1)} + \frac{\bar{\sigma}^4_1}{N^2_2(N_2-1)}}
\end{equation}

where $\bar{\mu}_i$ is the sample mean reward for distribution $i$, $\bar{\sigma}_i$ is the sample standard deviation for distribution $i$, and $N_i$ is the sample size  for distribution $i$. We compare the reward distributions of the top two actions, with highest expected reward, to determine the p-value. We ignore other actions because of the asymmetrical nature of an MCTS tree causing the worst actions to have very few rollouts.

%% file: sections/5_experiments.tex
\begin{figure}[t]
    \centering
    \subcaptionbox{\label{fig:grid_search}}{\includegraphics[height=3cm]{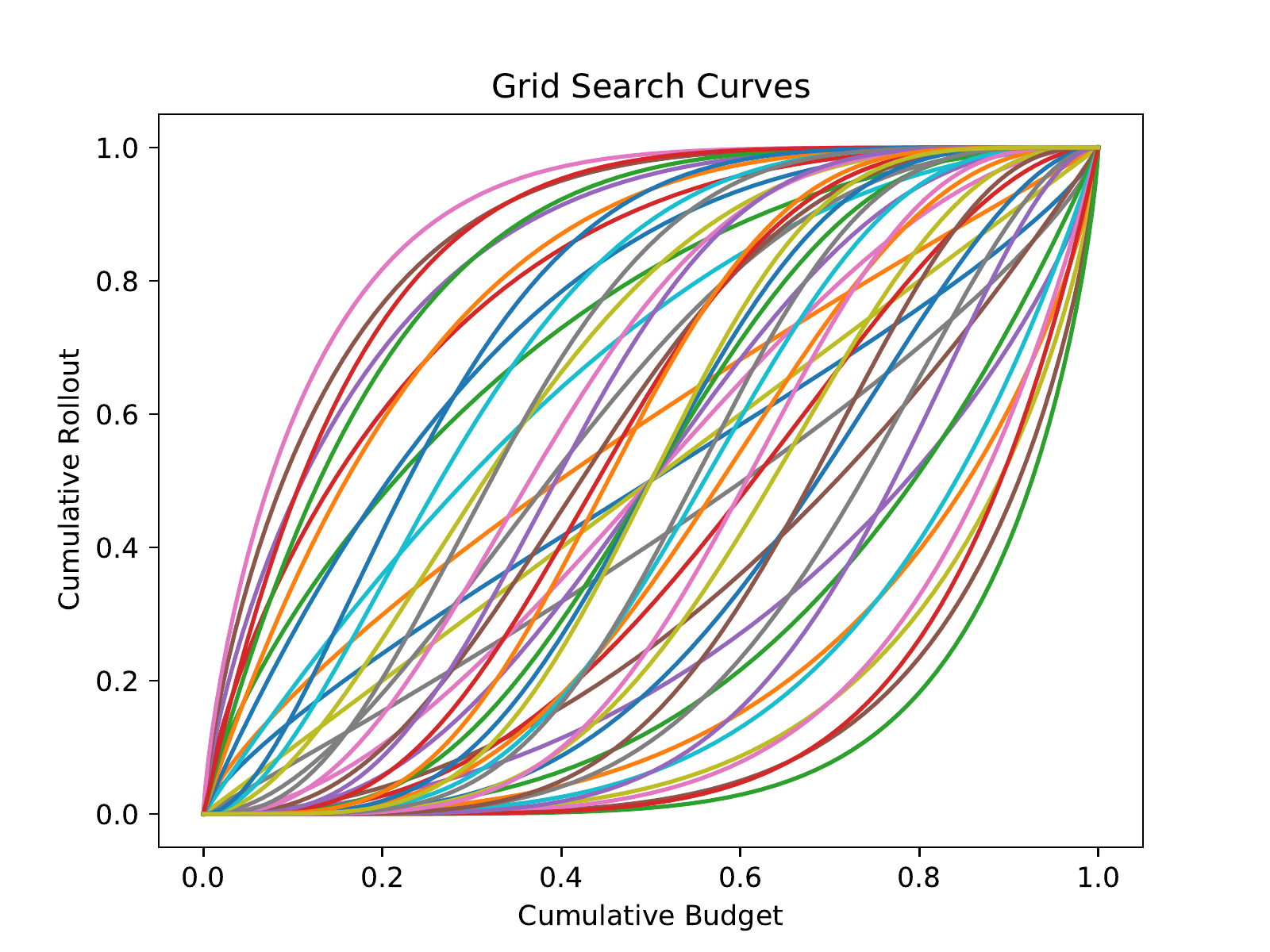}}
    \subcaptionbox{\label{fig:grid_search_results}}{\includegraphics[height=3cm]{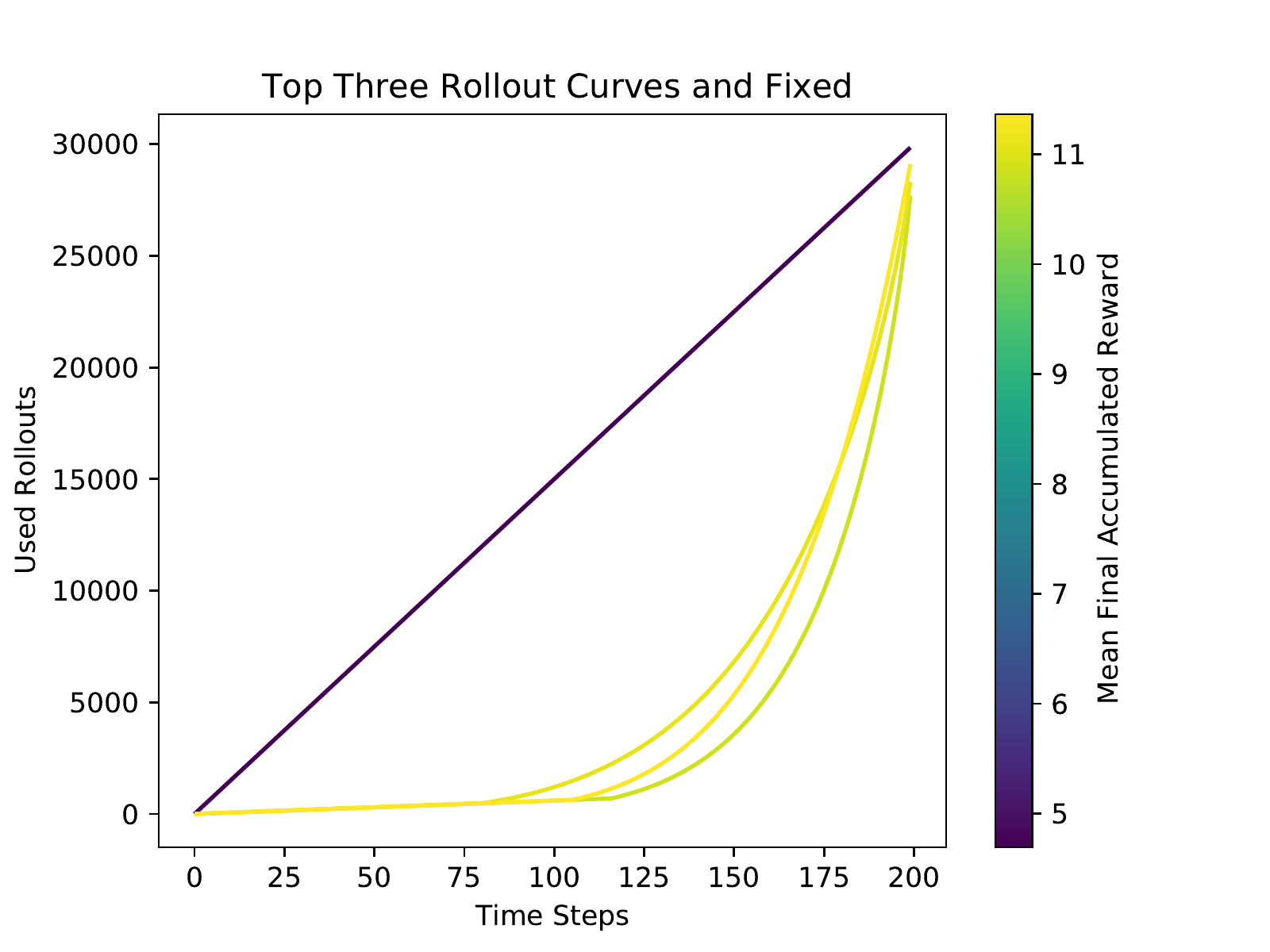}}
    \caption{Results from a grid search over possible rollout curves. \cref{fig:grid_search} presents the curves that were searched over. \cref{fig:grid_search_results} shows the three most optimal curves and a linear allocation curve, colored by their mean accumulated reward at the end of the episode.}
\vspace{-0.1in}
\end{figure}

\begin{figure*}[t] %
    \centering
    \subcaptionbox{\label{alc:sbo_fn}}{\includegraphics[width=0.19\textwidth]{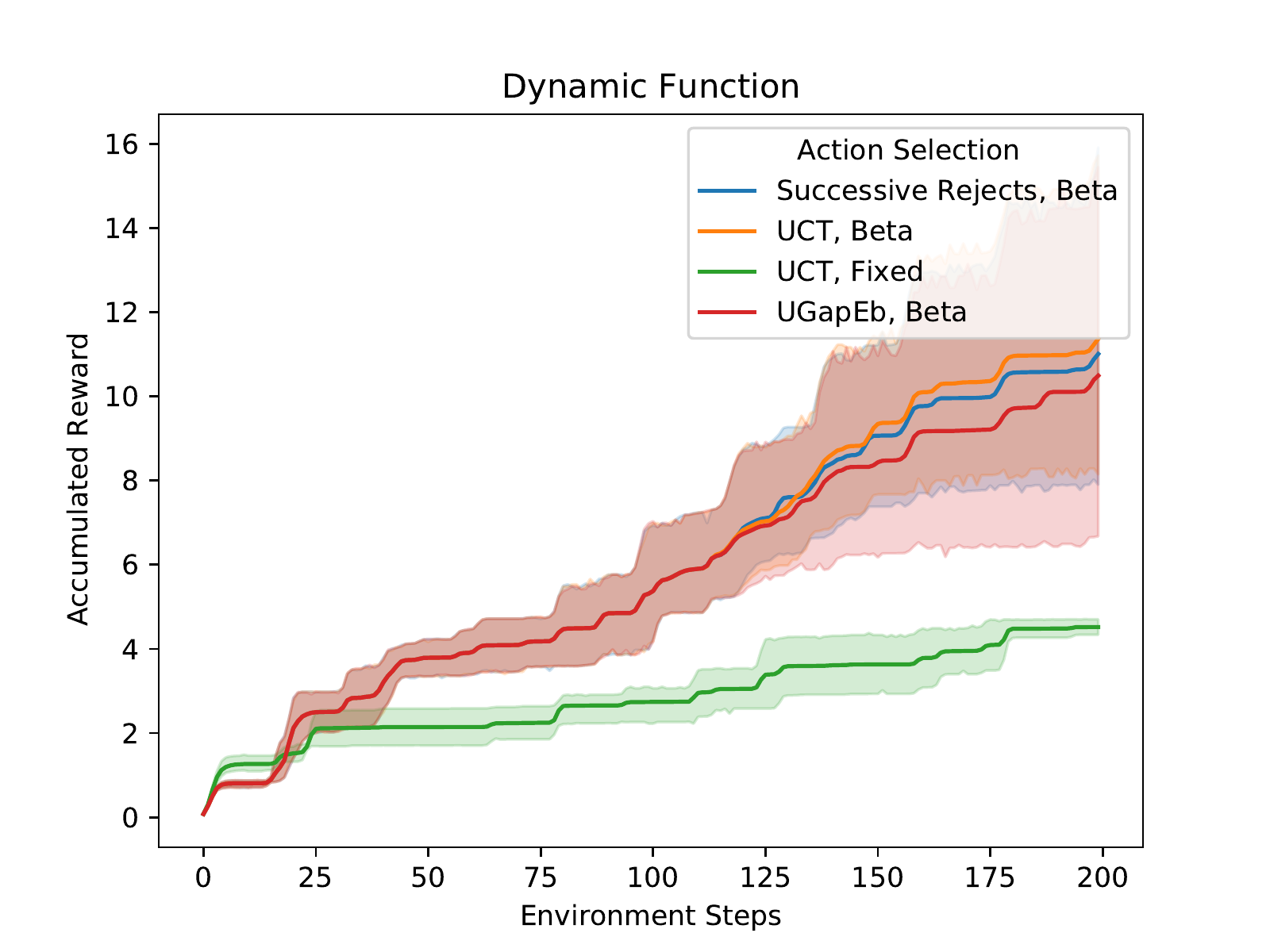}}
    \subcaptionbox{\label{alc:val1}}{\includegraphics[width=0.19\textwidth]{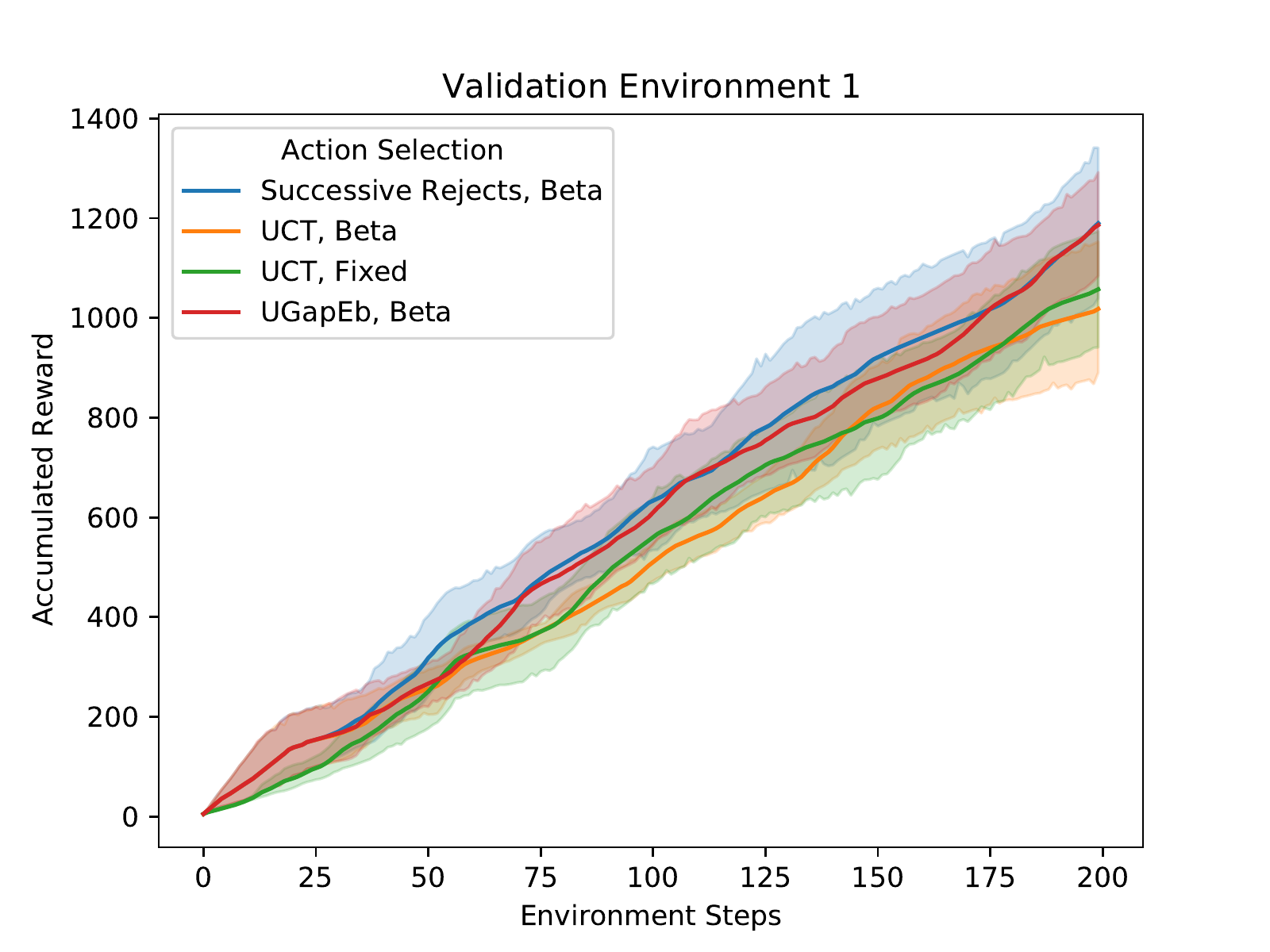}}
    \subcaptionbox{\label{alc:val2}}{\includegraphics[width=0.19\textwidth]{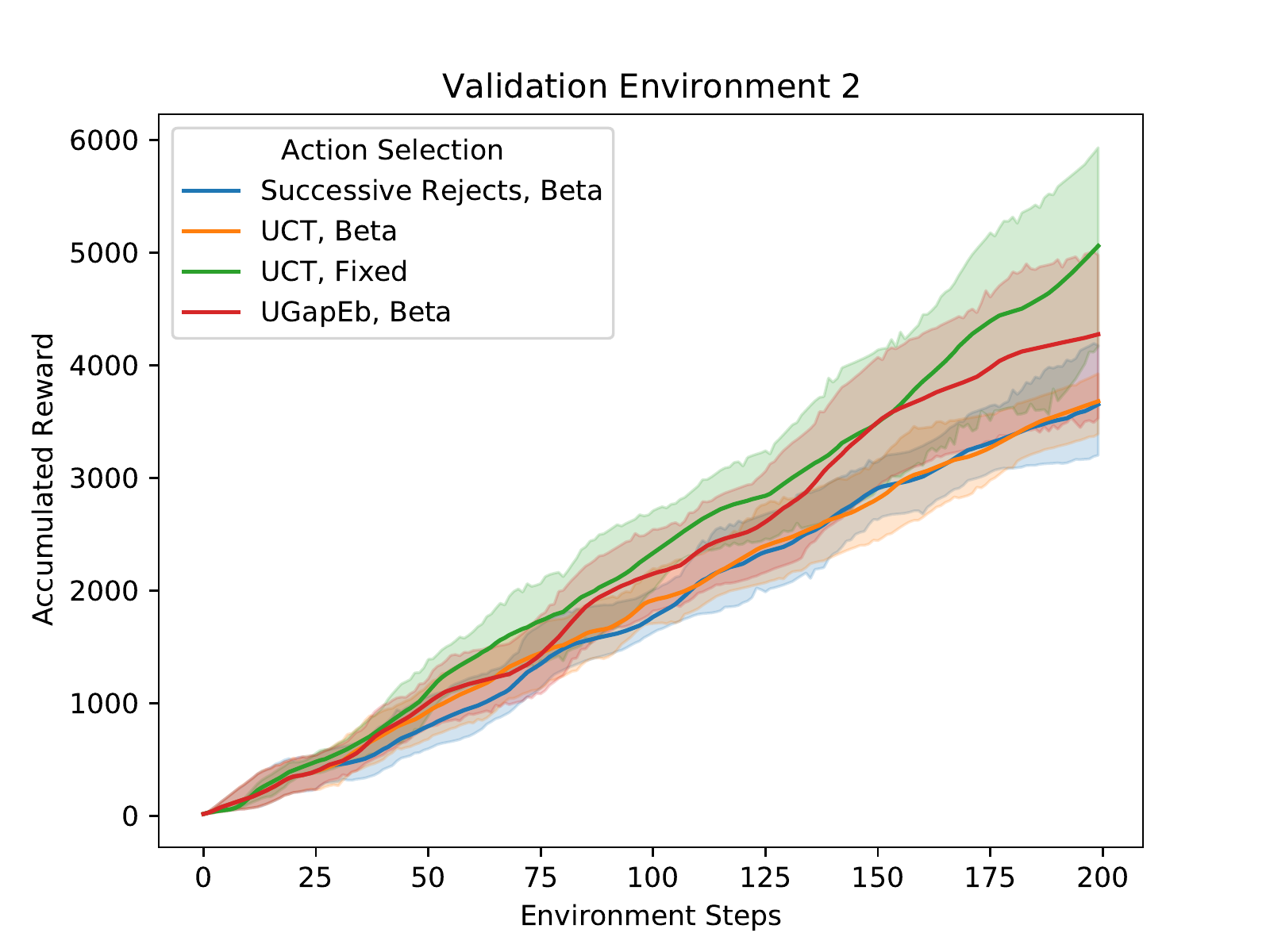}}
    \label{fig:exp_algorithm}
        \subcaptionbox{\label{fig:planCommit_reward}}{\includegraphics[width=0.19\textwidth]{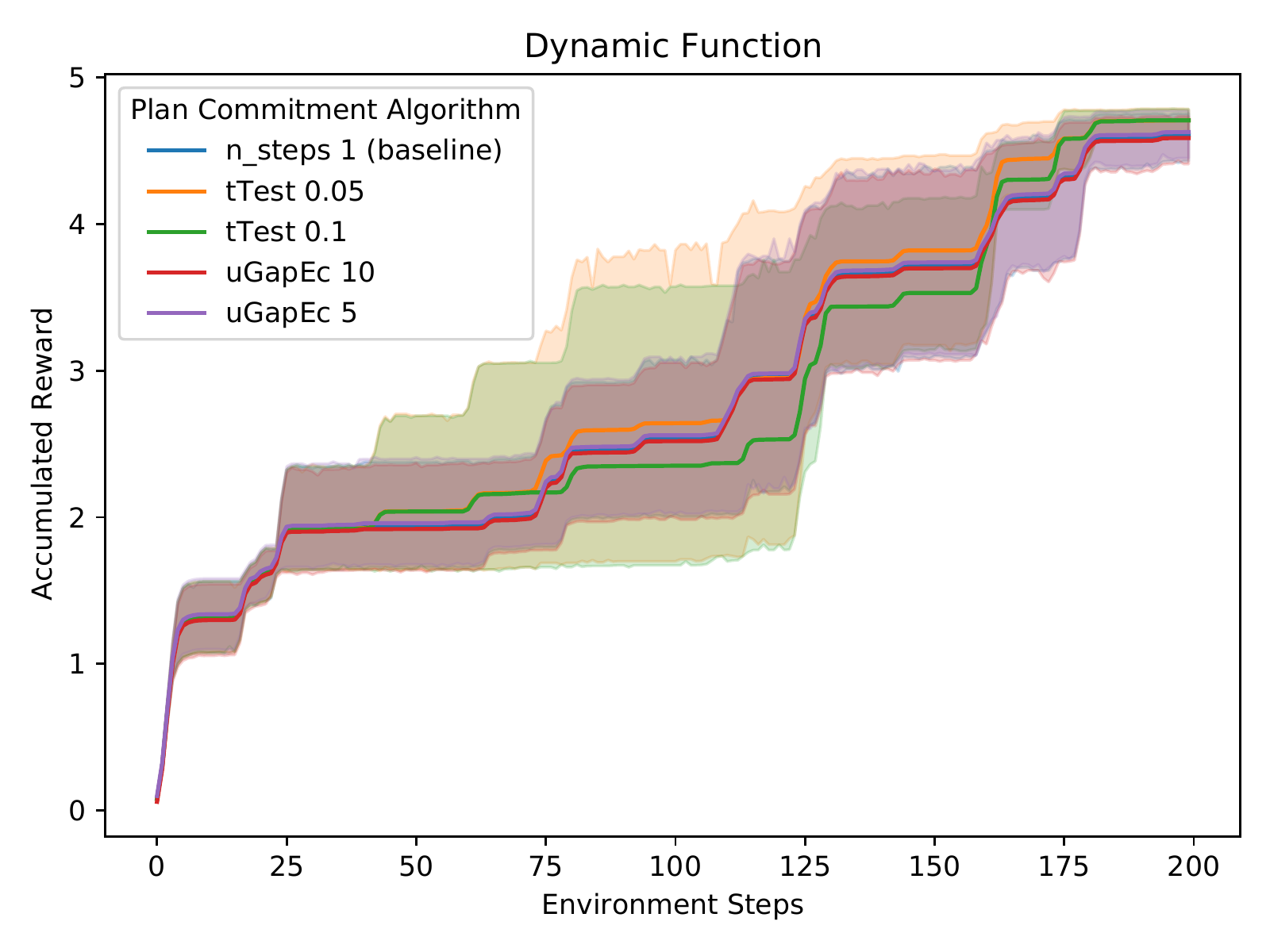}}
    \subcaptionbox{\label{fig:planCommit_rollouts}}{\includegraphics[width=0.19\textwidth]{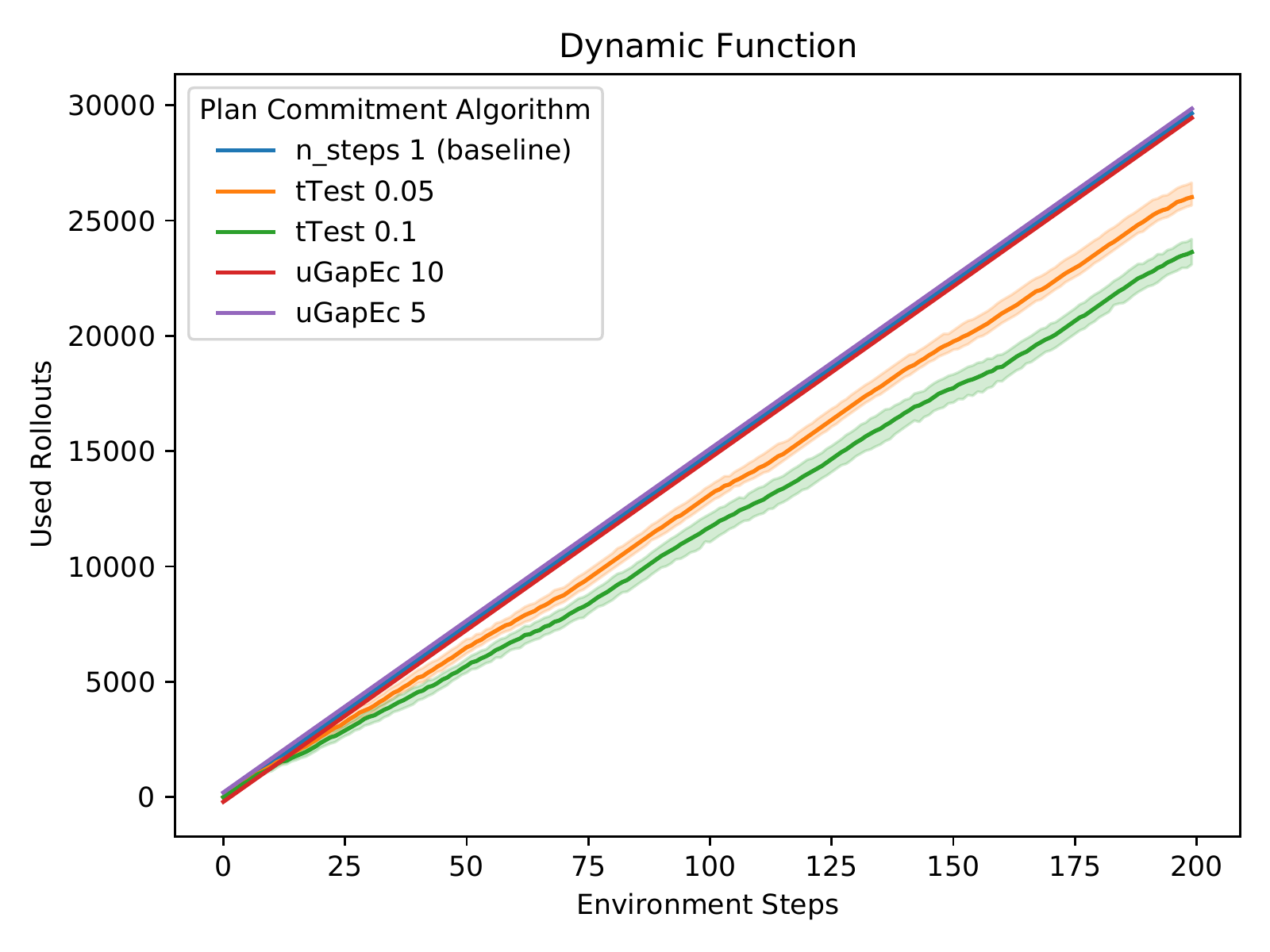}}
    \caption{Comparison of Exploration Algorithms (\subref{alc:sbo_fn}, \subref{alc:val1}, and \subref{alc:val2}) and Plan Commitment Algorithms (\subref{fig:planCommit_reward} and \subref{fig:planCommit_rollouts}). \textit{UCT, Fixed} is the baseline which evenly splits the rollouts at each step and uses the UCT exploration algorithm (the default for MCTS). Other results use a curved rollout allocation. For plan commitment, \cref{fig:planCommit_reward} shows the reward accumulation and \cref{fig:planCommit_rollouts} shows the number of rollouts used in all POMCP calls for the whole episode. A small offset value is added to methods which overlap.}
\vspace{-0.15in}
\end{figure*}

We assess the performance of our improvements on three environments.
The first environment is a test function for testing the effectiveness of sequential Bayesian Optimization using POMCP~\cite{Marchanta}.
We use a dynamic (time-varying) two dimensional function as the underlying ground truth for testing our individual improvements. It corresponds to a Gaussian curve circling a fixed point twelve times.
\begin{equation}\label{eq:dynamic_function}
  g(x,y,t) = e^{-(\frac{x - 2 - 1.5sin(24 \pi t)}{0.7})^2} e^{-(\frac{y - 2 - 1.5cos(24 \pi t)}{0.7})^2} 
\end{equation}
where $x \in [0,5], y \in [0,5], t \in [0,1]$. In this environment the agent starts out at the bottom-center of the time box and progresses towards the top, choosing actions in the x-y plane. A subsampled image of this environment (points below $g(x,y,t)<0.6$ removed for clarity) can be seen in \cref{fig:dynamic_function}.

\begin{figure}[t]
    \centering
   
    \subcaptionbox{\label{fig:env1}}{\includegraphics[height=2.7cm]{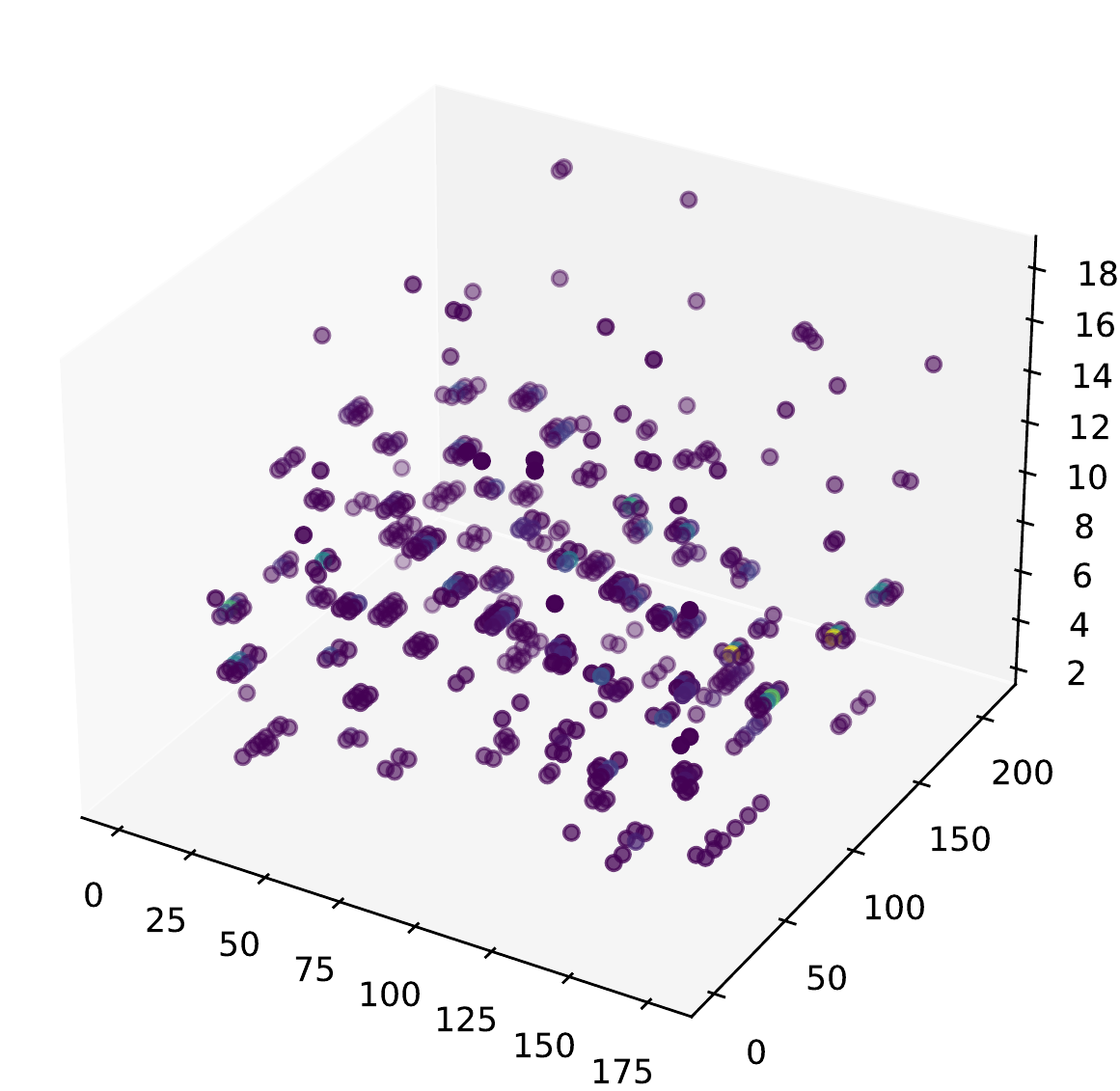}}
    \subcaptionbox{\label{fig:env1_baseline}}{\includegraphics[height=2.7cm]{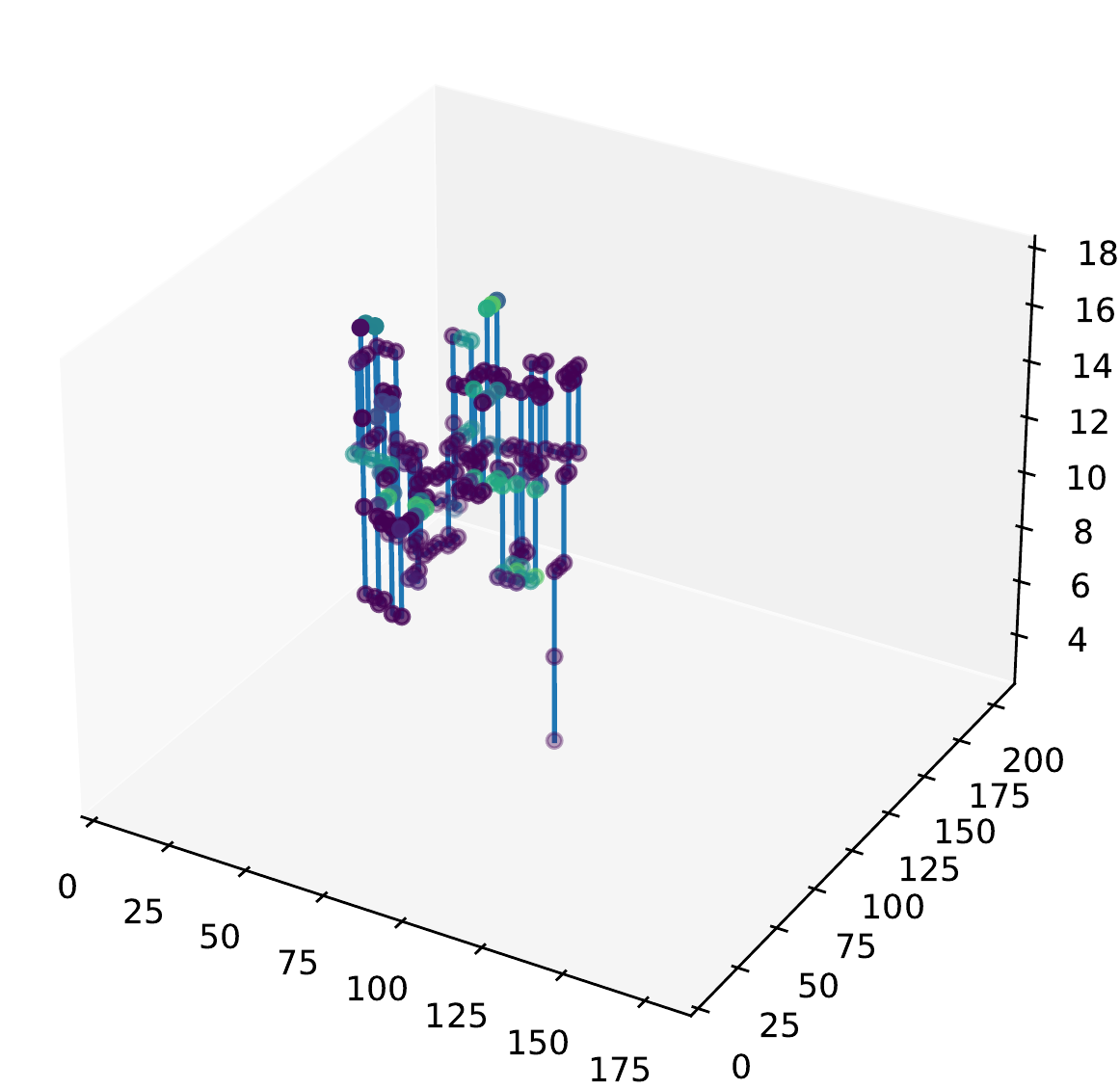}}
    \subcaptionbox{\label{fig:env1_ours}}{\includegraphics[height=2.7cm]{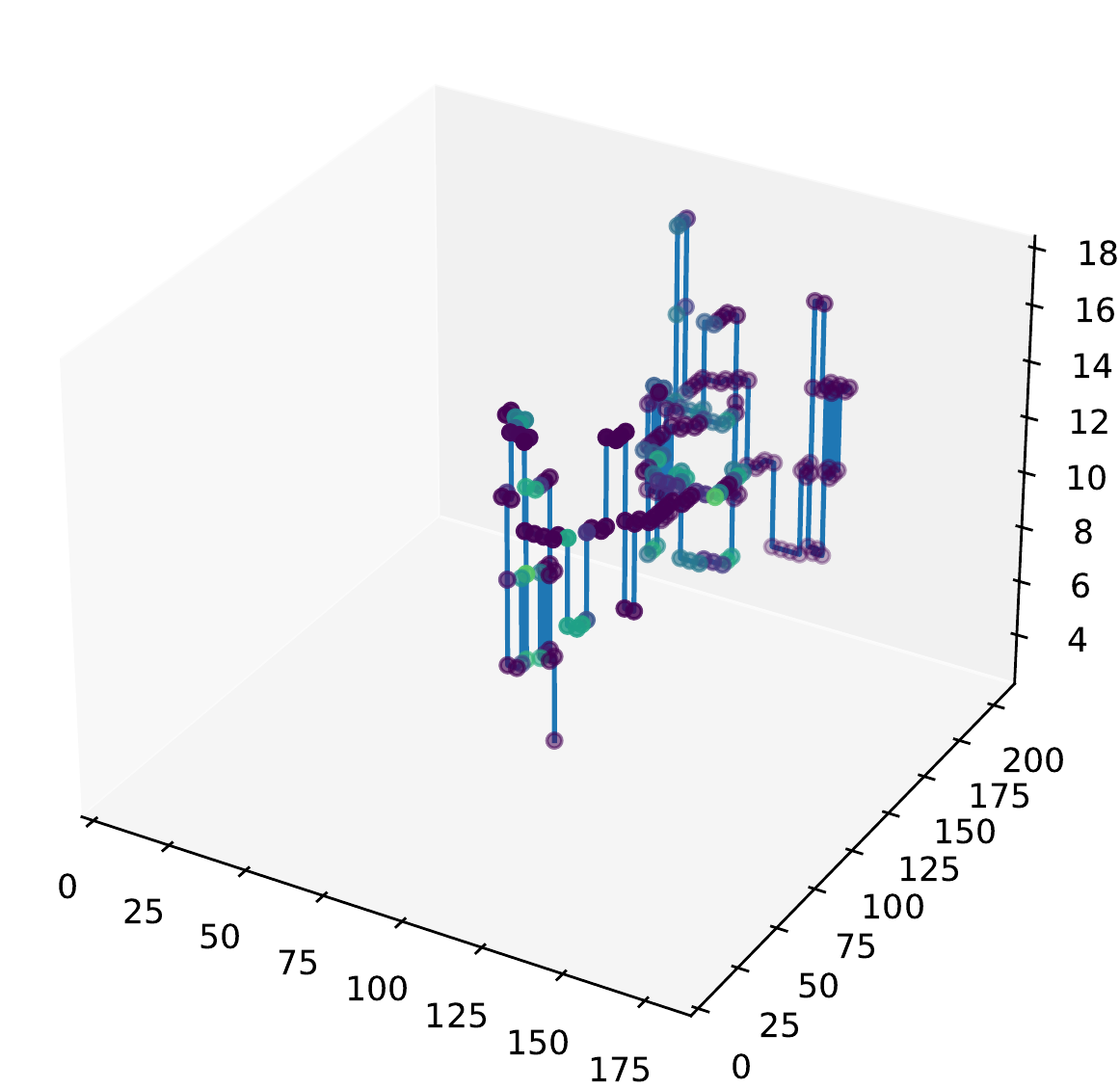}}
   
    \subcaptionbox{\label{fig:env2}}{\includegraphics[height=2.7cm]{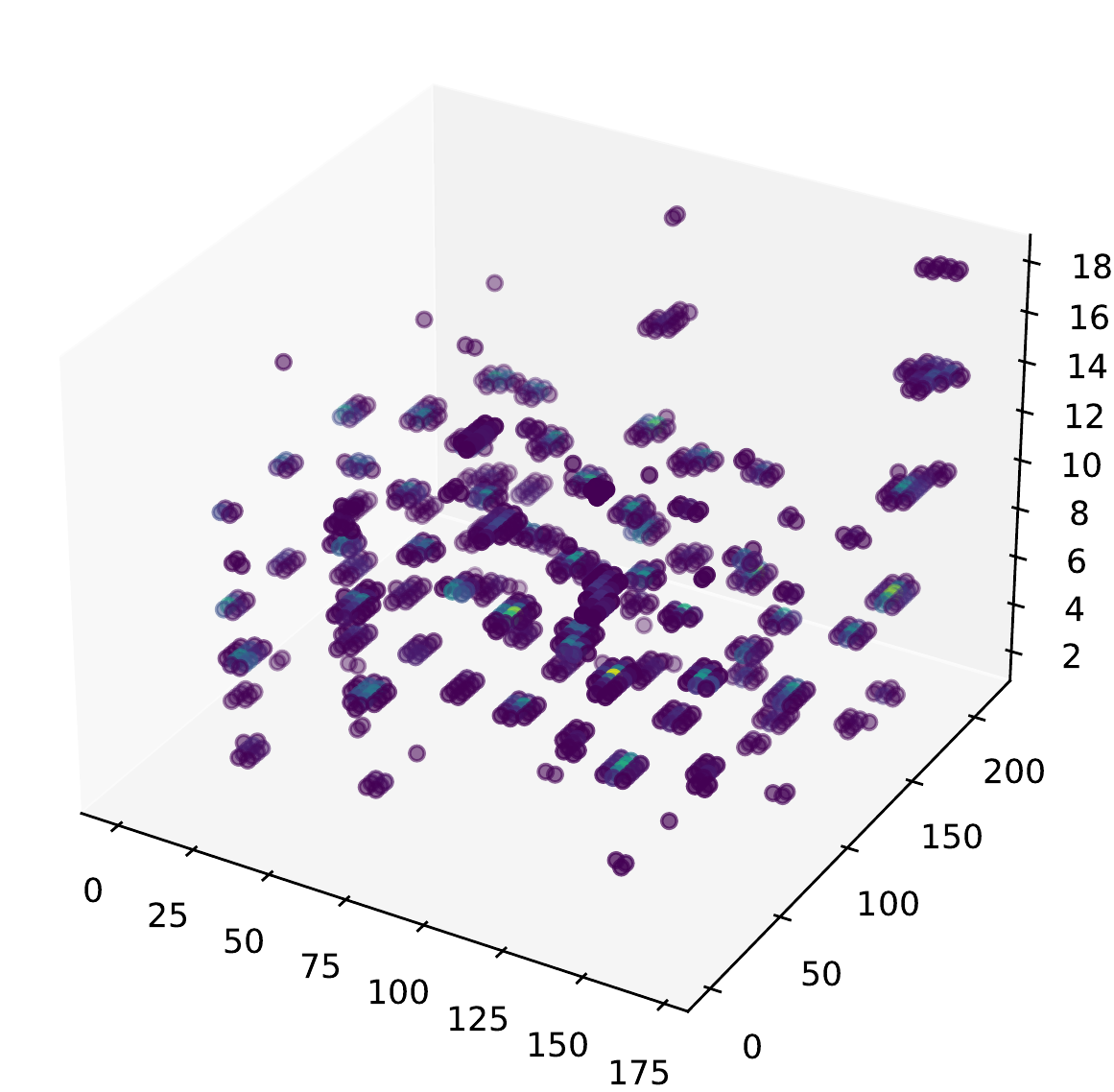}}
    \subcaptionbox{\label{fig:env2_baseline}}{\includegraphics[height=2.7cm]{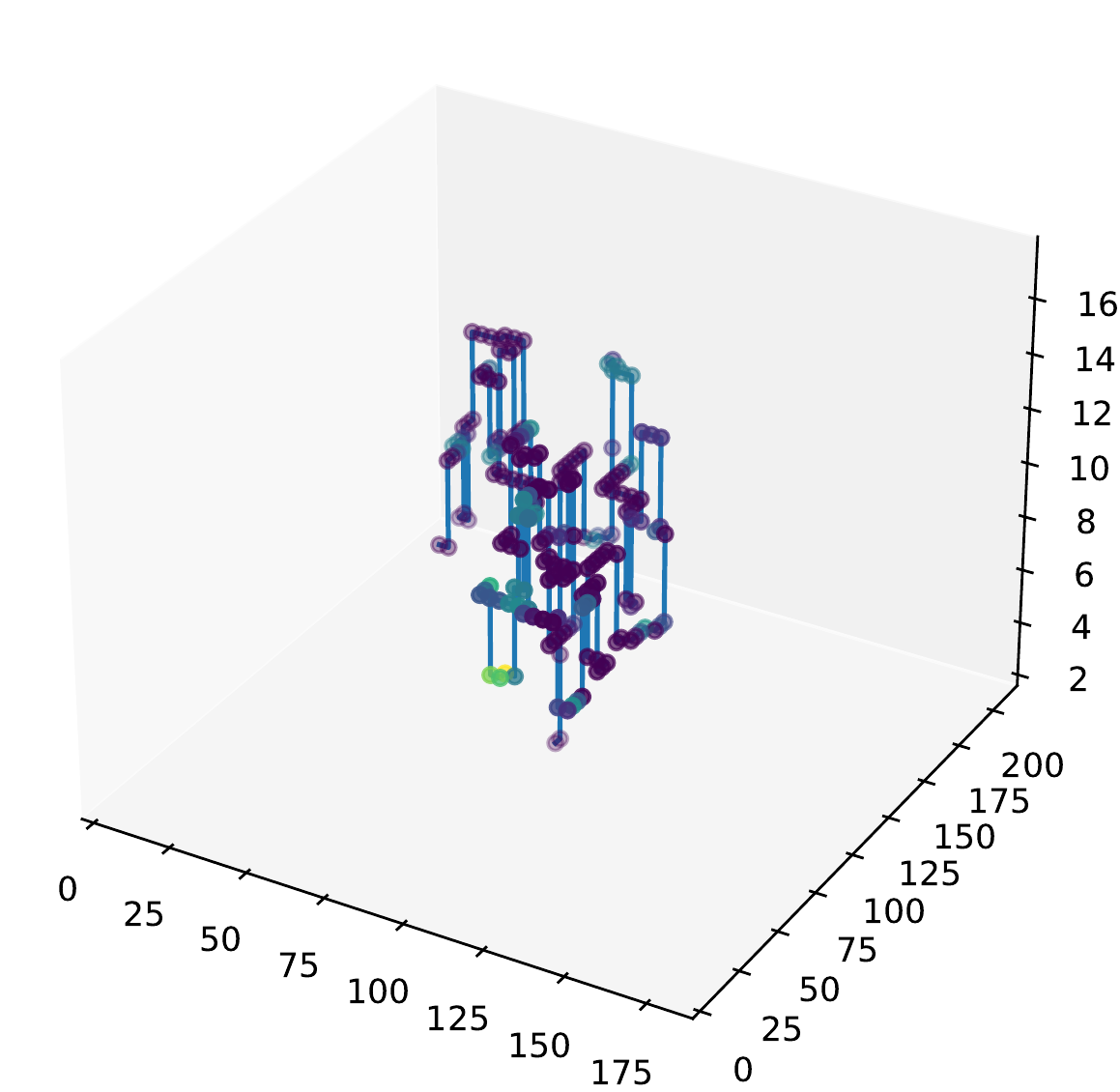}}
    \subcaptionbox{\label{fig:env2_ours}}{\includegraphics[height=2.7cm]{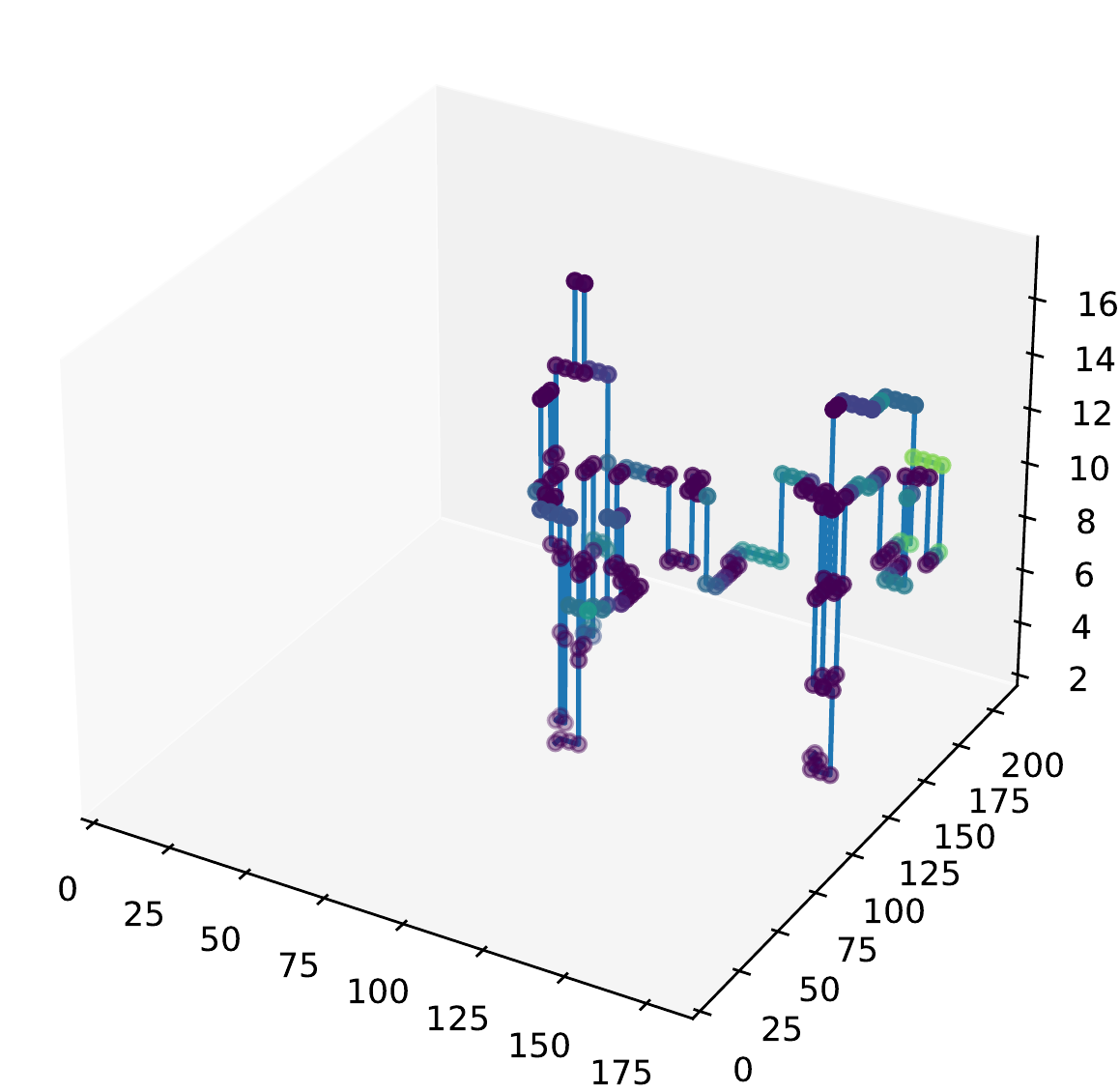}}

    \caption{\cref{fig:env1} shows a dataset collected with an underwater robot, and \cref{fig:env1_baseline} and \cref{fig:env1_ours} show example trajectories from a baseline implementation and our proposed implementation respectively. 
    \cref{fig:env2} shows another, more complex, dataset collected in the same location, \cref{fig:env2_baseline} and \cref{fig:env2_ours} show example trajectories from a baseline implementation and our proposed implementation respectively.}
\vspace{-0.25in}
\end{figure}

In the other two environments, called Validation Environment 1 (\cref{fig:env1}) and Validation Environment 2 (\cref{fig:env2}), we use chlorophyll concentration data collected from a YSI Ecomapper robot as input data.
These datasets are $186m$ by $210m$, by $15m$ deep.
We interpolate this with a Gaussian process to create the underlying function to estimate.
In these scenarios the robot travels $3m$ between each sample and can freely travel in any direction at any point.
The robot starts at the center of the environment at 0 depth.

For all environments, the agent is allowed to take 200 environment steps and is assumed to have a point motion model that can move to neighboring locations.
We use the objective function $\mu_x + c \sigma_x$ and use $c=10$ for the dynamic function and $c=100$ for the validation environments.
All experiments are run for five seeds each.

\subsection{Grid Search for Rollout Allocation}

To find the proper form of the rollout allocation and test the assertion that different parts of the POMDP planning process need different number of rollouts we perform a grid search over different curves that describe the rollout allocation.
For each curve, if it would allocate less than one rollout per action, we allow the planner to perform a single rollout per action.
We parameterize these curves by cumulative beta distributions because of their flexibility in representing many different kinds of curves.
These curves are parametrized by an $\alpha$ and $\beta$ parameter which determine the exact form of the curve.
We search over $\alpha=[.75,1,2,3,4,5,6]$ and $\beta=[.75,1,2,3,4,5,6]$.
These curves can be seen in \cref{fig:grid_search}.

The results of this experiment are shown in \cref{fig:grid_search_results}, which indicate that an exponential increase in the rollout allocations is desirable and a very flat curve is undesirable. 
We find that the best curve for the dynamic function to be $\alpha=6,\beta=1$.
We empirically find that this curve does worse on the validation environments, possibly due to overfitting, and that a curve with $\alpha=4,\beta=4$ works best. We use this for tests involving a curved rollout allocation with them.

\subsection{Comparison of Exploration Algorithms}
We test the effectiveness of alternative exploration algorithms to UCT and the interaction between the rollout allocation method and exploration algorithm. We test three exploration algorithms described in \cref{armSelection}: UGapEb, UCT, and Successive-Rejects on three environments. In \cref{alc:sbo_fn} all beta curve-based methods outperform the fixed method and all allocators work almost equally well, with UCT having a slight performance boost. In \cref{alc:val1}, UGapEb and Successive Rejects with curved rollout allocation perform approximately equally but out-perform UCT with both a fixed and curved rollout allocation. In \cref{alc:val2} all three curved allocators are out-performed by a fixed rollout allocation curve. This is likely because the rollout curve is poorly chosen for this environment due to not being chosen by grid search.
UGapEb outperforms all curved allocators by a significant margin.
\vspace{-0.1in}
\subsection{Comparison of Plan Commitment Algorithms}
We test the methods described in \cref{planCommitment} for determining how many steps to take once the MCTS tree is generated.
We test the unequal variances t-test and UGapEc methods with different parameters against a baseline method, which takes only the first action, across 5 seeds. 
\cref{fig:planCommit_reward} and \cref{fig:planCommit_rollouts} shows the comparison of all these combinations against the baseline. 
UGapEc and the baseline largely overlap because UGapEc cannot confidently predict whether the chosen action is the best action with such few rollouts and a larger epsilon does not make sense for the scale of the rewards.
We believe that UGapEc may be of use for complex environments where the agent cannot make strong assumptions about the underlying reward distributions and many more rollouts will be required for the POMCP algorithm.
The unequal variances t-test performs the best amongst the options. 
Within the t-test parameters, the p-value of 0.1 requires slightly fewer rollouts than a p-value of 0.05 for similar reward. 
However, choosing 0.1 implies riskier behavior which can have a negative effect in complex environments and real-world datasets, such as our validation environments. 
Hence, we choose the unequal variance t-test with p = 0.05 as our best choice for plan commitment.
\cref{fig:planCommit_reward} shows the accumulated reward for a trajectory execution between the baseline and our choice. 
In \cref{fig:planCommit_rollouts} it is clear that each of the algorithms take vastly different amounts of rollouts for obtaining this result.
Hence, we see that the plan commitment t-test algorithm helps to significantly reduce the rollouts needed to solve the adaptive sampling problem. %

\begin{figure}[t]
    \centering
    \subcaptionbox{\label{fig:combo_expt_sbo_reward}}{\includegraphics[width=4.25cm]{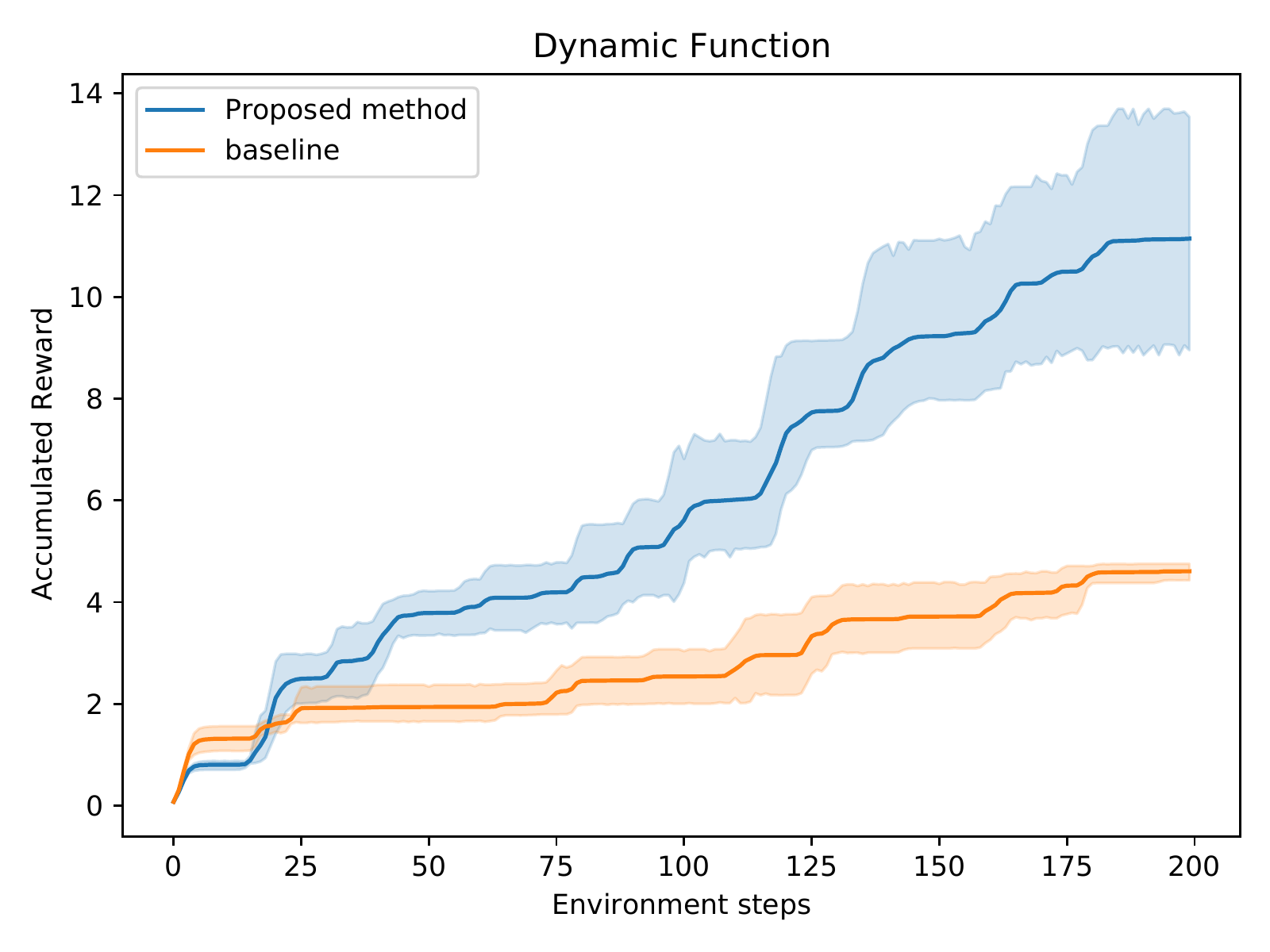}}
    \subcaptionbox{\label{fig:combo_expt_sbo_rollouts}}{\includegraphics[width=4.25cm]{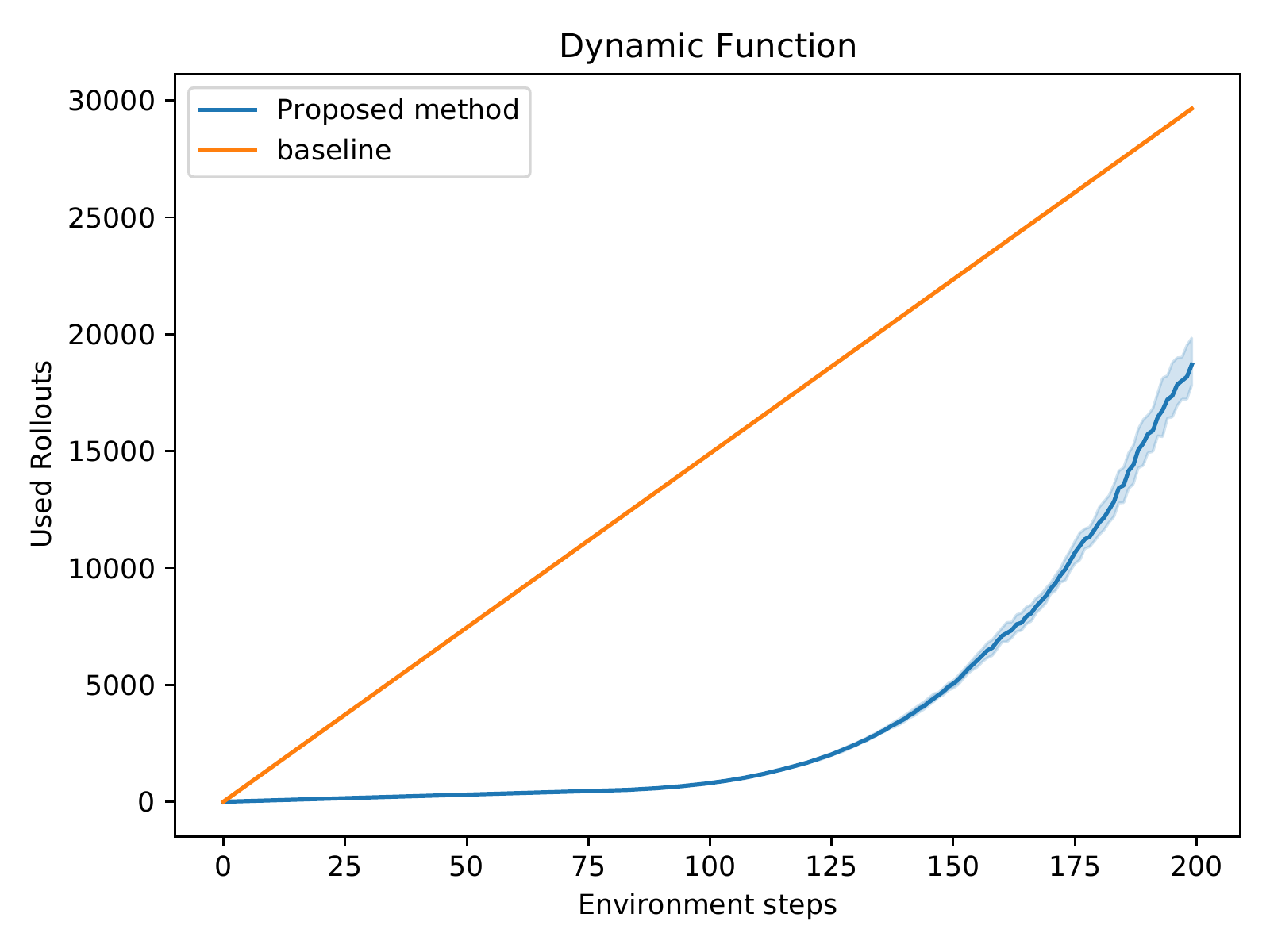}}
    \subcaptionbox{\label{fig:combo_expt_env1_reward}}{\includegraphics[width=4.25cm]{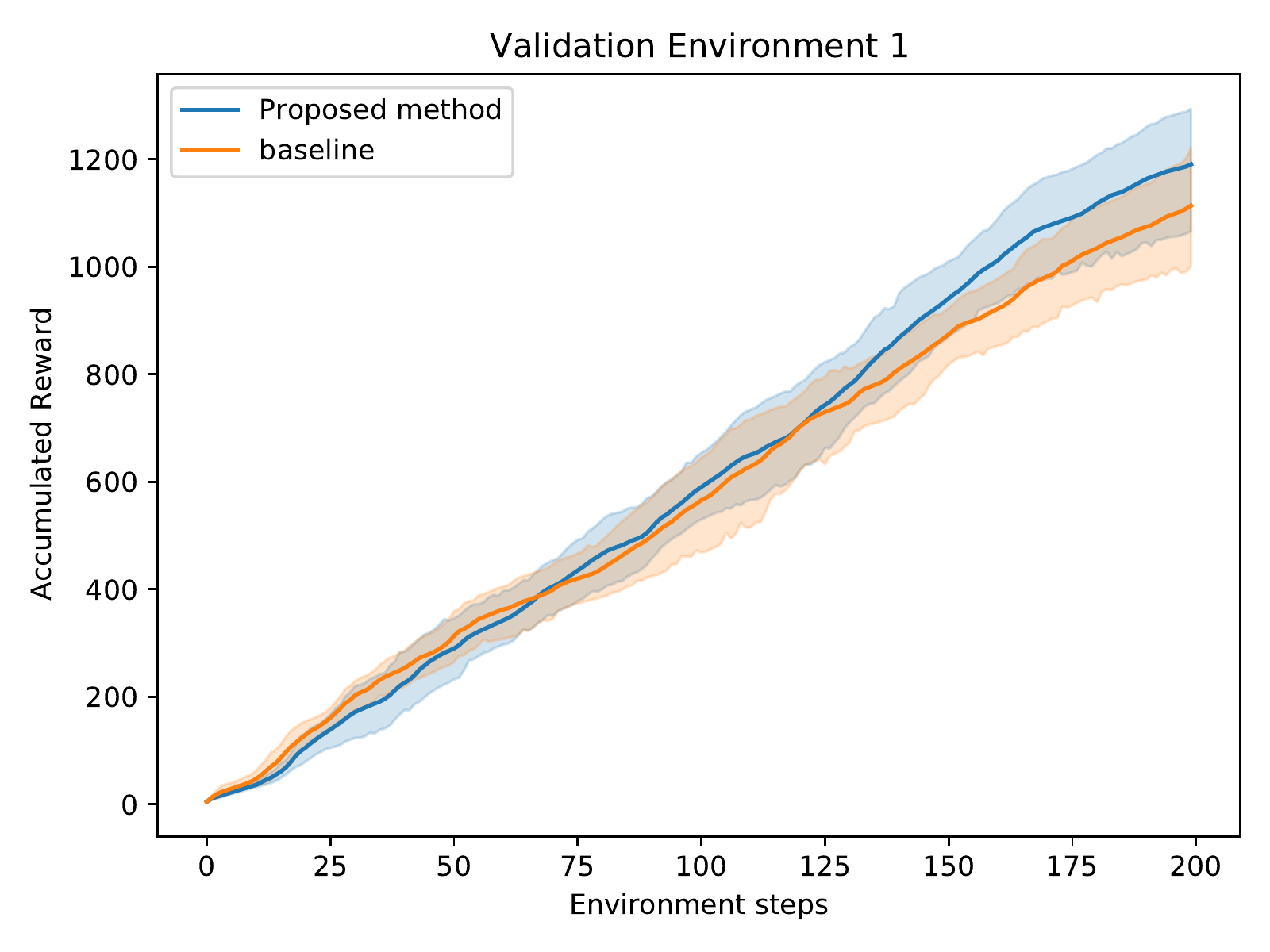}}
    \subcaptionbox{\label{fig:combo_expt_env1_rollouts}}{\includegraphics[width=4.25cm]{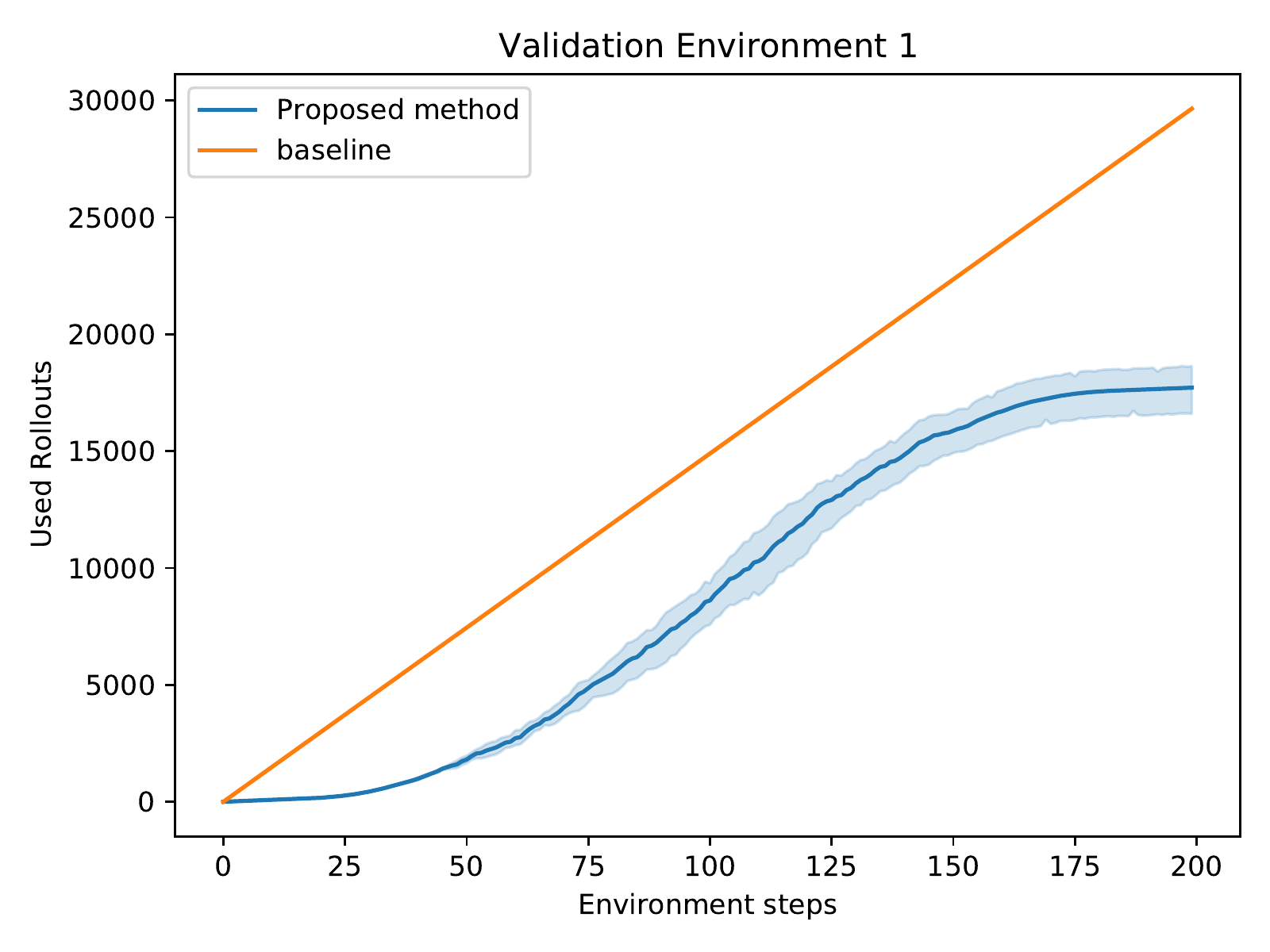}}
    \subcaptionbox{\label{fig:combo_expt_env2_reward}}{\includegraphics[width=4.25cm]{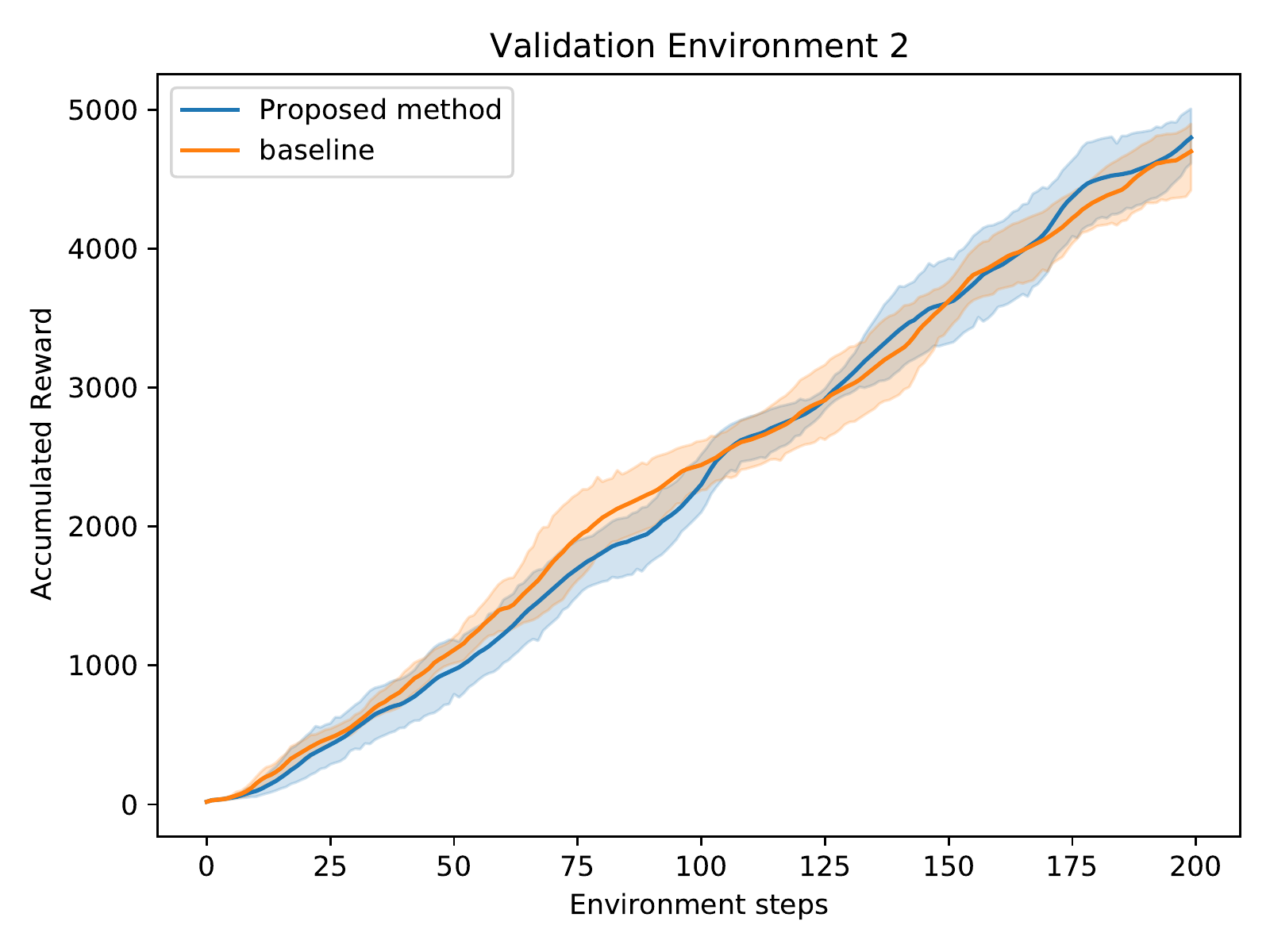}}
    \subcaptionbox{\label{fig:combo_expt_env2_rollouts}}{\includegraphics[width=4.25cm]{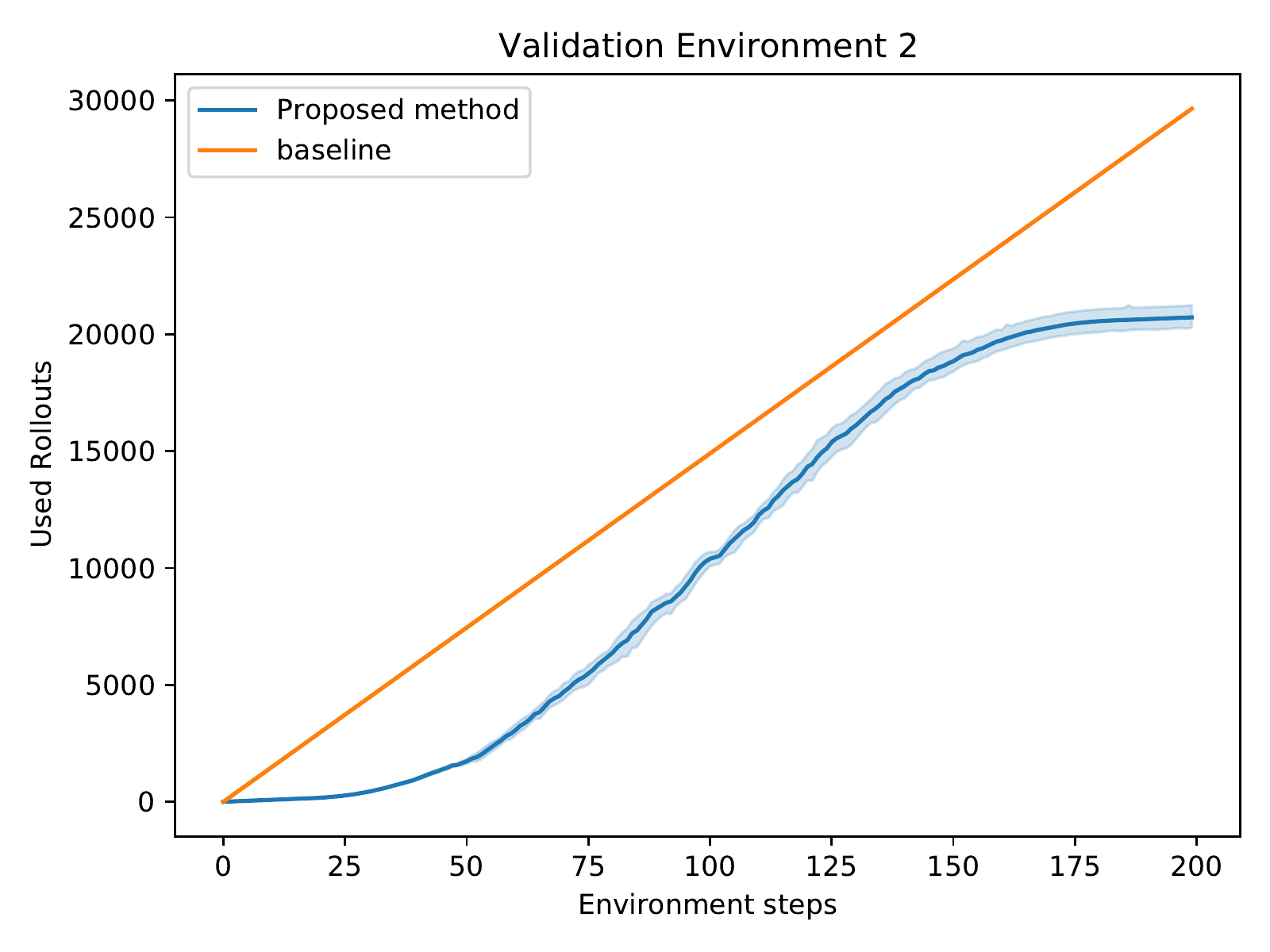}}
    \caption{Comparison of the combined proposed improvements (Proposed Method) against the baseline for all environments. \cref{fig:combo_expt_sbo_reward} and \cref{fig:combo_expt_sbo_rollouts} are the reward and number of rollouts used by the agent in the dynamic function environment. \cref{fig:combo_expt_env1_reward} and \cref{fig:combo_expt_env1_rollouts} are the reward and number of rollouts used by the agent in Validation Environment 1, \cref{fig:combo_expt_env2_reward} and \cref{fig:combo_expt_env2_rollouts} are the reward and number of rollouts used by the agent in Validation Environment 2.}
    \label{fig:combo_expt}
\vspace{-0.25in}
\end{figure}
\subsection{Comparison of Baseline with Combined Improvements}
\cref{fig:combo_expt} shows the combined effect of all improvements from the preceding experiments: using a curved rollout allocation, using the UGapEb exploration algorithm and using the t-test plan commitment algorithm. 
We compare against a baseline which uses an equal number of rollouts at each step, uses the UCT exploration algorithm and takes only one action before replanning.
We compare our method and this baseline for each environment. 
\cref{fig:combo_expt_sbo_reward} and \cref{fig:combo_expt_sbo_rollouts} show that the combined features achieve a much higher reward in fewer rollouts on the dynamic environment.
\cref{fig:combo_expt_env1_reward} and \cref{fig:combo_expt_env2_rollouts} show that again the agent receives a higher reward in many fewer rollouts than the baseline method.
\cref{fig:combo_expt_env2_reward} and \cref{fig:combo_expt_env2_rollouts} indicate that our method is comparable to the baseline in terms of reward but achieves this reward in fewer rollouts.

%% file: sections/6_conclusion.tex
We present improvements for online adaptive sampling with a Monte Carlo-based POMDP solver which uses specific knowledge of the adaptive sampling problem structure. This allows the agent to estimate a non-parametric function by taking samples of the underlying phenomenon such as the concentration of chlorophyll in a body of water.

First, we show that by changing the amount of rollouts that are allocated to more heavily favor later stages in planning, a better overall model of the environment can be created.
We believe this is due to later stages having more information to plan on and therefore able to develop better and longer plans.
We show that searching for an optimal curve can lead to high performance increases and that reasonable curves chosen can lead to increased performance. Second, we show that the agent's total reward can increase by changing the action exploration algorithm to one that explicitly incorporates knowledge of the number of rollouts allocated for each planning step.
This works with the rollout allocation to improve selection when few rollouts are allocated. 
We also show that by modifying the amount of steps the agent takes from a planning tree, the overall planning can be made more efficient. We show a statistical test can be used to determine if an action can be confidently determined to be the best action. With this test we are able to reduce the number of rollouts needed to reach a comparable accumulated reward. Finally, we show that these improvements are synergistic and when used together can greatly improve the planning over a fixed-step, optimal exploration, fixed-rollout allocation planner.

%% file: sections/8_appendix.tex
\subsection{Comparison of time saving to baseline}
Because our proposed method reduces the number of rollouts, the time to compute a plan is reduced. 
Additionally, rollouts from later environment steps are cheaper to compute because the remaining budget is lower. 
This causes our method to finish the episode faster because it allocates more rollouts to later environment steps.
These two effects combine to produce a saving in wall-clock time for the entire episode, as shown in \cref{tab:wallclock}.
Experiments were run on a server with 2 Intel$^{\textregistered}$ Xeon Gold processors and 256GB RAM.

\begin{table}[h!]
    \centering
    \begin{tabular}{p{1.1cm} p{1.4cm} p{1.7cm} p{1.7cm} }
    Method & Dynamic Function & Validation \mbox{Environment} 1 & Validation \mbox{Environment} 2 \\
    \hline\hline
    Baseline & 2061.84 & 2687.73 & 3542.72 \\
    \hline
    Proposed Method & 1371.77 & 2497.36  & 3120.90 \\
    \end{tabular}
    \caption{Wall-clock time (seconds) required to complete five episodes.}
    \label{tab:wallclock}
\vspace{-0.25in}
\end{table}

\subsection{Visualization of Dynamic Function}
\begin{figure}[h!]
    \centering
    \includegraphics[width=\columnwidth]{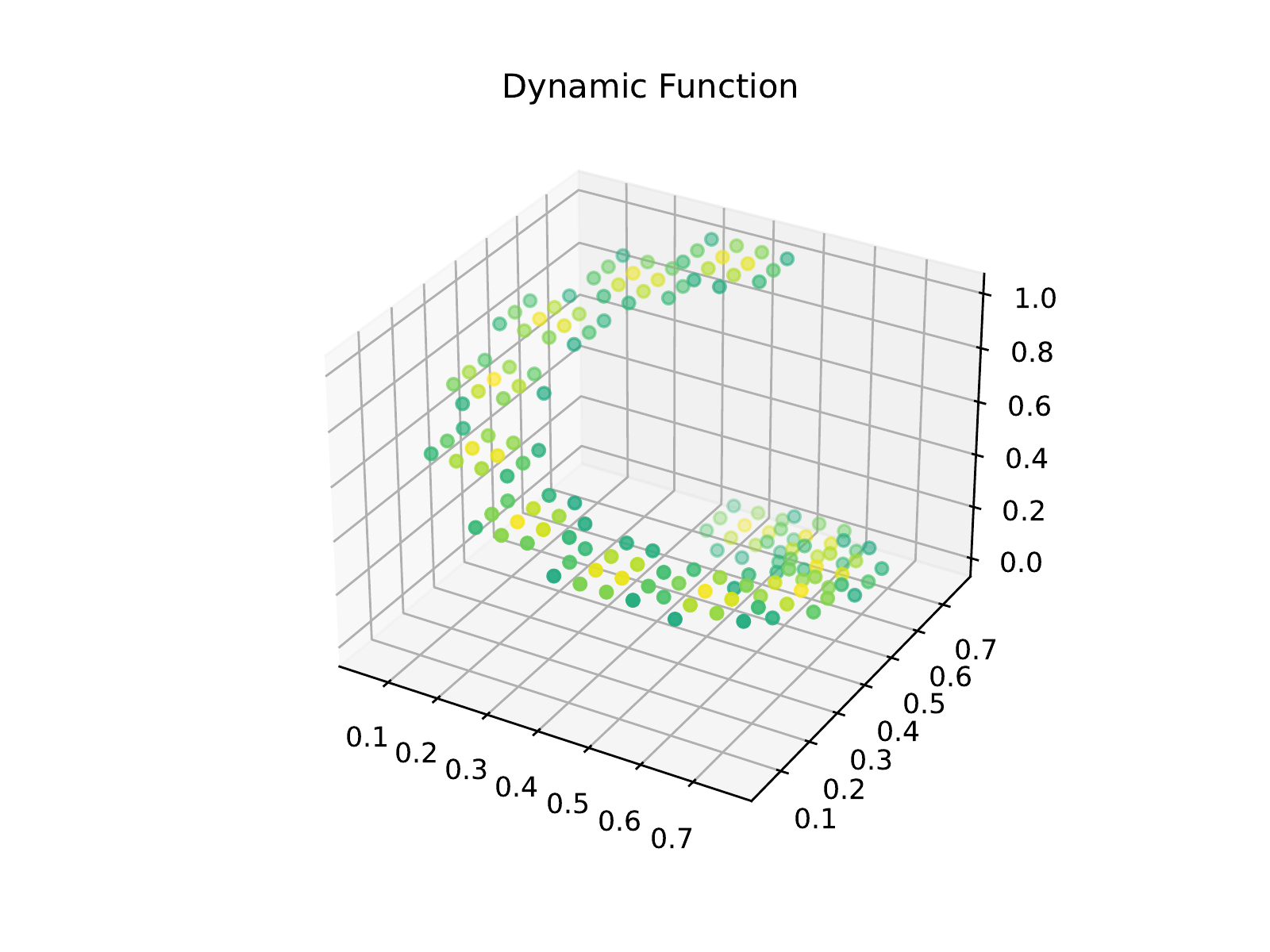}
    \caption{The dynamic (time-varying) two dimensional function used for testing the effectiveness of our method, described by \cref{eq:dynamic_function}. Note that this is a subsampled image, only showing values above $g(x,y,t) \geq 0.6$, for clarity.}
    \label{fig:dynamic_function}
\end{figure}

We present a visualization of the dynamic function used for testing. 
The function can be seen in \cref{fig:dynamic_function}.

%% file: sections/7_future_work.tex
Currently the rollout allocation algorithm either requires an expensive grid search or an a-priori guess. 
We would like to determine the correct rollout curve online and adapt to the information the agent has seen.

Future directions may also include environments where the underlying reward distributions are farther away from a Gaussian distribution. 
In this case, methods like uGapEc or other MAB methods that do not make assumptions about the underlying distributions may perform better.